\documentclass{ieeetj}

\usepackage[english]{babel}
\usepackage{textcomp}
\usepackage{amsmath}
\usepackage{amssymb}
\usepackage{booktabs}
\usepackage{caption}
\usepackage{subfigure}
\usepackage{multirow}
\usepackage{amsfonts}
\usepackage{graphicx}
\usepackage{breakurl}
\usepackage{enumerate}
\usepackage{tabularx}
\usepackage{algorithmic}
\usepackage{algorithm}
\usepackage{bm}
\usepackage{subcaption}
\usepackage{enumerate}
\usepackage{adjustbox}
\usepackage{xcolor}
\usepackage{siunitx}
\usepackage{booktabs}
\usepackage{mdframed}
\usepackage{adjustbox}
\usepackage{color}
\usepackage{xurl}
\usepackage{makecell}
\usepackage{cite}
\makeatletter
\let\NAT@parse\undefined
\makeatother
\usepackage{hyperref}
\usepackage{relsize}
\usepackage{float}
\usepackage{pifont}
\hypersetup{hidelinks=true}

\DeclareMathOperator*{\argmin}{arg\,min}

\def\BibTeX{{\rm B\kern-.05em{\sc i\kern-.025em b}\kern-.08em T\kern-.1667em\lower.7ex\hbox{E}\kern-.125emX}}
\AtBeginDocument{\definecolor{tmlcncolor}{cmyk}{0.93,0.59,0.15,0.02}\definecolor{NavyBlue}{RGB}{0,86,125}}

\def\OJlogo{}
\def\seclogo{}

\begin{document}
% \receiveddate{XX Month, XXXX}
% \reviseddate{XX Month, XXXX}
% \accepteddate{XX Month, XXXX}
% \publisheddate{XX Month, XXXX}
% \currentdate{XX Month, XXXX}
% \doiinfo{XXXX.2022.1234567}
\markboth{NOVA: Navigation via Object-Centric Visual Autonomy for High-Speed Target Tracking in Unstructured GPS-Denied Environments}{\textit{Saviolo et al.}}

\title{NOVA: Navigation via Object-Centric Visual Autonomy for High-Speed Target Tracking in Unstructured GPS-Denied Environments}
% \title{Target Tracking beyond the Lab:\\Model Predictive Control with Visual Feedback for High-speed Navigation in Unknown GPS-Denied Environments}
% \title{High-Speed Target Tracking of UAV Using Visual Feedback}

\author{Alessandro Saviolo and Giuseppe Loianno}
\affil{New York University, New York, NY 11201 USA}
\corresp{Corresponding author: Alessandro Saviolo (email: alessandro.saviolo@nyu.edu).}
\authornote{``This work was supported by the NSF CAREER Award 2145277, DARPA YFA Grant D22AP00156-00, Qualcomm Research, Nokia, and NYU Wireless.''}

% 150-250 words
\begin{abstract}
Autonomous aerial target tracking in unstructured and GPS-denied environments remains a fundamental challenge in robotics. Many existing methods rely on motion capture systems, pre-mapped scenes, or feature-based localization to ensure safety and control, limiting their deployment in real-world conditions. We introduce \textbf{NOVA}, a fully onboard, object-centric framework that enables robust target tracking and collision-aware navigation using only a stereo camera and an IMU.
Rather than constructing a global map or relying on absolute localization, NOVA formulates perception, estimation, and control entirely in the target’s reference frame. A tightly integrated stack combines a lightweight object detector with stereo depth completion, followed by histogram-based filtering to infer robust target distances under occlusion and noise. These measurements feed a visual-inertial state estimator that recovers the full 6-DoF pose of the robot relative to the target.
A nonlinear model predictive controller (NMPC) plans dynamically feasible trajectories in the target frame. To ensure safety, high-order control barrier functions (CBFs) are constructed online from a compact set of high-risk collision points extracted from depth, enabling real-time obstacle avoidance without maps or dense representations.
We validate NOVA across challenging real-world scenarios, including urban mazes, forest trails, and repeated transitions through buildings with intermittent GPS loss and severe lighting changes that disrupt feature-based localization.
Each experiment is repeated multiple times under similar conditions to assess resilience, showing consistent and reliable performance. 
NOVA achieves agile target following at speeds exceeding $50~\si{km/h}$. 
These results show that high-speed, vision-based tracking is possible in the wild using only onboard sensing, with no reliance on external localization or assumptions on the environment structure. \textbf{Video}: \url{https://youtu.be/UpbcSKQrHfk}
\end{abstract}


\maketitle

\begin{figure*}[t]
    \centering
    \includegraphics[width=\textwidth, trim=0 270 300 0, clip]{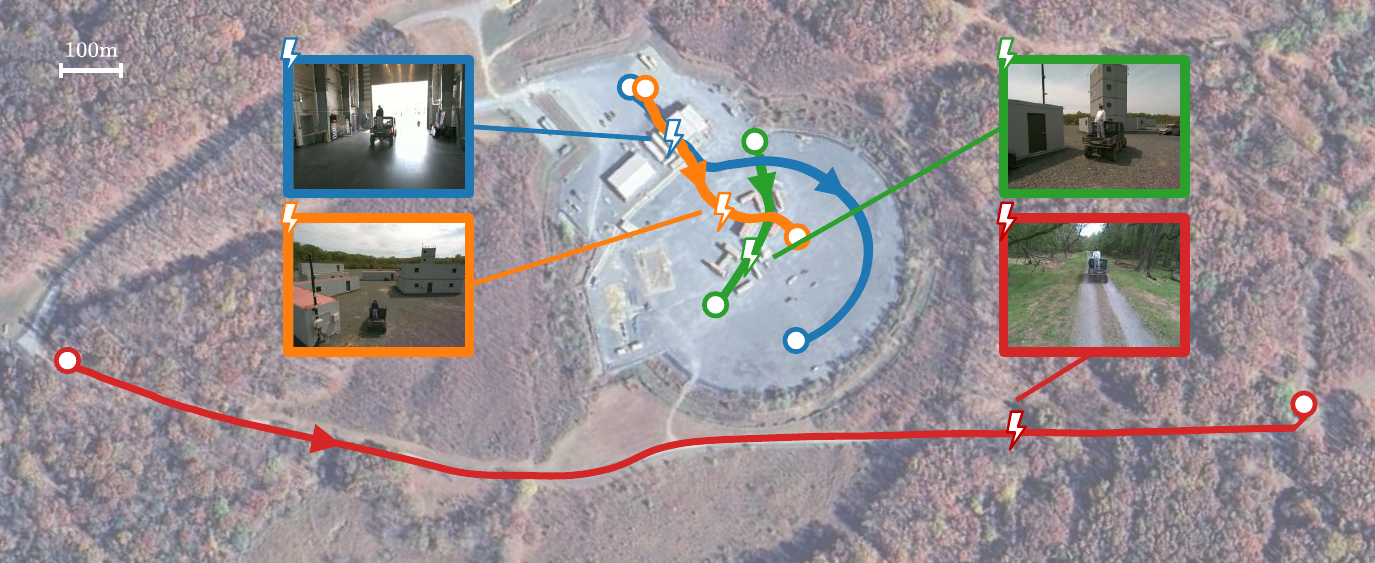}
    \caption{
    \textbf{Satellite imagery of representative NOVA flight missions in real-world environments.}
    Each overlaid UAV trajectory illustrates a distinct experimental scenario designed to test specific failure modes in visual-inertial target tracking and control. 
    \textcolor{red}{\textbf{Red:}} High-speed tracking in a forest trail over $1~\si{km}$, with target speeds exceeding $50~\si{km/h}$ and motion blur. 
    \textcolor{blue}{\textbf{Blue:}} Indoor--outdoor transition with complete GPS loss, exposure collapse, and perceptual ambiguity due to structural occlusions and distractors. 
    \textcolor{orange}{\textbf{Orange:}} Elevated tracking from a $10~\si{m}$ offset, challenging stereo depth perception, feature consistency, and planning geometry. 
    \textcolor{green}{\textbf{Green:}} Urban container maze with narrow corridors, cluttered turns, and minimal texture. 
    Insets show raw onboard RGB frames during each flight. NOVA maintains real-time target lock, depth estimation, obstacle avoidance, and closed-loop control using only a stereo camera and IMU. \textit{GPS is not used by NOVA and appears here only for visualization of flight paths.}
    }
    \label{fig:outdoor_paths}
\end{figure*}

\section{INTRODUCTION}
\IEEEPARstart{T}{arget} tracking is the ability of an autonomous system to detect, follow, and predict the motion of an object of interest over time. For Unmanned Aerial Vehicles (UAVs), this capability is critical in dynamic and unstructured environments where pre-mapped routes or fixed waypoints are insufficient. In such scenarios, robust tracking allows the UAV to remain persistently coupled to targets such as people, vehicles, or infrastructure, despite challenges like scene changes, occlusions, or unpredictable target motion.

This capability is key to diverse applications. In search and rescue, tracking allows UAVs to follow moving individuals through unstructured terrain, maintaining proximity without relying on maps or GPS~\cite{wu2021vision}. In human–robot interaction, it enables drones to interpret and respond to human motion in real time, facilitating adaptive behavior~\cite{xu2023visual, hu2022vision}. In inspection and surveillance, tracking ensures continuous observation of mobile assets, even under occlusion or changing viewpoints~\cite{gu2018multiple, bashir2022closed, huang2021online}.
Moreover, tracking is foundational in multi-robot systems, where relative positioning between agents is essential. UAVs in leader–follower formations or aerial swarms must track teammates to maintain coordinated motion~\cite{quan2022formation, di2021multi, miki2019uav}. In autonomous landing, a UAV must descend onto a moving platform, requiring precise tracking~\cite{niu2021vision, demirhan2020development, vlantis2015quadrotor}.

Visual tracking methods for UAVs traditionally fall into two categories: Image-Based Visual Servoing (IBVS) and Position-Based Visual Servoing (PBVS)~\cite{chaumette2007visual, hutchinson1996tutorial, cho2019sampling, oliva2022towards, raj2016appearance}. IBVS operates directly on image-plane measurements, allowing low-latency control and robustness to calibration errors. However, its effectiveness is limited by the camera's field of view and sensitivity to target occlusion or rapid motion. PBVS instead reconstructs the target's 3D position, typically in a globally referenced frame, and uses it to guide control. While PBVS mitigates many of IBVS’s limitations, it introduces a new dependency: accurate consistent global localization.

In GPS-denied settings such as forests, urban mazes, indoor spaces, or extraterrestrial terrain, global localization typically relies on Visual-Inertial Odometry (VIO)~\cite{geneva2020openvins, qin2018vins, scaramuzza2019visual} or Simultaneous Localization and Mapping (SLAM)~\cite{labbe2019rtab, aguilar2017visual, mur2015orb}, which estimate the robot’s pose by tracking visual features across image sequences. These methods assume sufficient texture, consistent lighting, and a relatively static environment to maintain reliable feature correspondences.
However, unstructured environments often violate these assumptions due to cluttered geometry, dynamic elements, ambiguous semantics, and degraded visibility. As a result, feature-based localization quickly becomes unreliable~\cite{wang2018improving}.

A notable example of this failure was NASA’s Ingenuity Mars Helicopter~\cite{tzanetos2022ingenuity, balaram2021ingenuity, withrow2020advanced}, which suffered a navigation failure while flying over smooth, featureless sand ripples~\cite{grip2022flying}. Without distinctive landmarks, its visual-inertial system produced erroneous velocity estimates, leading to a crash 20 seconds after takeoff. Had Ingenuity been able to localize relative to a known object in the scene, such as the Perseverance rover, it might have maintained stable flight and avoided the crash. This example highlights a broader insight: when a known and observable target is available, it can serve as a reliable frame of reference than the surrounding terrain.

This paper introduces \textbf{NOVA}, a unified framework for \textit{Navigation via Object-centric Visual Autonomy}. NOVA enables real-time aerial target tracking and obstacle-aware navigation using only onboard sensing, with no dependence on GPS, external maps, or motion capture. Unlike traditional VIO or SLAM-based methods that require globally consistent localization, NOVA formulates perception, estimation, and control directly in the target’s frame of reference. This object-centric approach enables tracking in environments where localization is unreliable or unavailable.

At the core of NOVA is a tightly integrated perception-to-control stack that couples a custom object detector, depth completion module, and visual-inertial state estimator. The object detector operates in real time on low-resolution inputs using an adaptive zoom strategy to maintain tracking at long ranges. Depth estimates are computed using a fusion of stereo and monocular cues, then processed via histogram-based filtering to produce robust target distance estimates, even under occlusion or noise. From this, a full 6-DoF state of the robot relative to the target is inferred in real time. Crucially, this perception system is designed to operate under realistic sensing conditions, such as degraded lighting, partial occlusions, and unstructured terrain.

These target-centric estimates are passed to a Nonlinear Model Predictive Controller (NMPC) that optimizes dynamically feasible trajectories directly in the target’s reference frame. To ensure safe operation in cluttered environments, obstacle avoidance is enforced online using high-order Control Barrier Functions (CBFs), derived from a compact set of high-risk collision points identified in the depth map. This eliminates the need for dense mapping or environmental priors, enabling real-time, map-free collision avoidance.

We validate NOVA across a suite of challenging real-world experiments, illustrated in Figure~\ref{fig:outdoor_paths}. Each scenario targets a specific failure mode common to traditional visual tracking pipelines. In the forest trail mission, the UAV follows a moving target at speeds exceeding $50~\si{km/h}$ over more than $1~\si{km}$, inducing severe motion blur and visual degradation from airborne dust. The building transition scenario involves abrupt lighting changes and full GPS loss as the target moves from an open parking lot into an enclosed space and back outdoors. In the height-offset trial, the UAV maintains a vertical height of $10~\si{m}$, challenging feature matching due to reduced parallax and resolution.

Across all experiments, NOVA maintains stable target lock, plans feasible trajectories, and avoids obstacles in real time. To evaluate repeatability, each mission is conducted multiple times under similar conditions, with consistent performance observed throughout. All sensing and computation are performed fully onboard using only a stereo camera and an IMU. Unlike prior methods that rely on motion capture, artificial landmarks, or pre-built maps, NOVA operates autonomously in unstructured environments with no external infrastructure. 
\textit{To the best of our knowledge, this is the first demonstration of real-time, high-speed object-centric navigation that operates fully onboard and in-the-wild, without external infrastructure, explicit mapping, or dependency on structured priors.}

\section{RELATED WORKS}

\subsection{Traditional Visual Servoing for Target Tracking}
Visual servoing provides a well-established framework for controlling robotic motion with respect to a target, typically categorized into IBVS and PBVS~\cite{chaumette2007visual, hutchinson1996tutorial, cho2019sampling, oliva2022towards, raj2016appearance}. In IBVS, control commands are computed by minimizing the error between observed and desired image-space features. These features are extracted from the camera view and mapped to robot motion using an interaction matrix that links image variation to movement. This method is efficient and effective for tasks where the target remains in view and image-space motion corresponds smoothly to inputs.

PBVS, on the other hand, estimates the full 3D pose of the target using stereo vision, depth sensors, or geometric priors, and computes control actions in Euclidean space. It aims to minimize the spatial error between the current and desired poses of the robot relative to the target. This spatial reasoning improves robustness when the target leaves the field of view. However, PBVS depends on accurate global pose estimates of both the robot and the target, requiring consistent global localization across frames. In GPS-denied or visually degraded environments, this dependency often becomes a critical point of failure.

While both paradigms are effective in structured, static environments~\cite{zhang2021robust, guo2020image, serra2016landing, zheng2016image, thomas2015visual}, their assumptions often break down in aerial target tracking. UAVs operate under rapid motion and wide viewpoint changes, which distort target appearance and invalidate interaction models. During aggressive maneuvers, UAVs experience frequent occlusions, motion blur, and target scale variation, causing visual features to become unreliable. IBVS fails when the target exits the frame, undergoes large appearance shifts, or when image-space motion no longer correlates predictably with control. PBVS struggles in GPS-denied environments or texture-sparse scenes where global pose estimation is not available or stable.

These challenges are further exacerbated by the limited computational resources available onboard UAVs and the stringent real-time requirements for flight control. In practice, neither classical IBVS nor PBVS alone can reliably handle the demands of high-speed, long-range target tracking in unstructured, dynamic environments.

\subsection{Data-driven Aerial Target Tracking}
Recent approaches in aerial target tracking have leveraged the power of deep learning to enhance perception and decision-making under challenging conditions. One prominent direction is the use of end-to-end learning, where a neural network directly maps raw sensory inputs to low-level control actions. These models, often trained via reinforcement learning or imitation learning, have demonstrated fast reaction times and adaptability to perceptual noise~\cite{ma2023target, dionigi2023exploring, mao2022auv, zhang2018coarse}. However, this tight coupling of perception and control sacrifices interpretability, and poses significant challenges in generalization to unseen environments. Moreover, training such models requires extensive data collected in diverse, unstructured settings, an effort that is not only costly but also difficult to scale. Unstructured outdoor environments, in particular, exhibit extreme variability in appearance, geometry, and semantics, making it impractical to fully cover their distribution during training~\cite{min2024autonomous}.

To address these limitations, a more modular paradigm has emerged where perception and control are treated as distinct but tightly integrated subsystems. In this setting, the perception module is responsible for detecting and localizing the target, while the control module translates this information into safe and efficient motion commands. Extensive research has been devoted to each side of the pipeline. On the perception front, the computer vision community has produced a range of approaches for object detection and tracking, from compact architectures optimized for real-time inference on embedded platforms~\cite{khanam2024yolov11, zhao2024detrs, zhang2022bytetrack, aharon2022bot}, to large-scale foundation models capable of open-set recognition and zero-shot generalization~\cite{cheng2024yolo, wang2025yoloe, ravi2024sam, liu2024grounding}. Techniques including temporal attention, depth prediction, and visual transformers have improved robustness to occlusion and visual degradation.

Alongside advances in perception, control design has emerged as a key focus in enabling robust aerial target following. While model-free approaches such as reinforcement learning have demonstrated flexibility and generalization~\cite{saviolo2024unifying, hu2025finite, kumar2024adaptive, leomanni2024quadrotor}, they often struggle to provide consistent safety or performance guarantees, particularly under distribution shifts or in dynamic settings. In contrast, model-based methods, especially those based on NMPC~\cite{jiang2024perception, altan2020model, gonzalez2010robust, sugie2010model}, offer a structured way to embed physical constraints, safety margins, and task-specific objectives directly into the control policy. By explicitly optimizing trajectories over a finite horizon, NMPC enables predictive and context-aware behaviors, which are critical for maintaining target visibility and avoiding obstacles during high-speed, reactive flight~\cite{saviolo2024reactive}.

This separation of perception and control not only improves system modularity and interpretability, but also allows for leveraging the complementary strengths of data-driven learning and analytical planning. However, the efficacy of such hybrid systems depends critically on how well the two modules are coupled.

\begin{figure*}[hbt!]
    \centering
    \includegraphics[width=\linewidth, trim=0 300 0 0, clip]{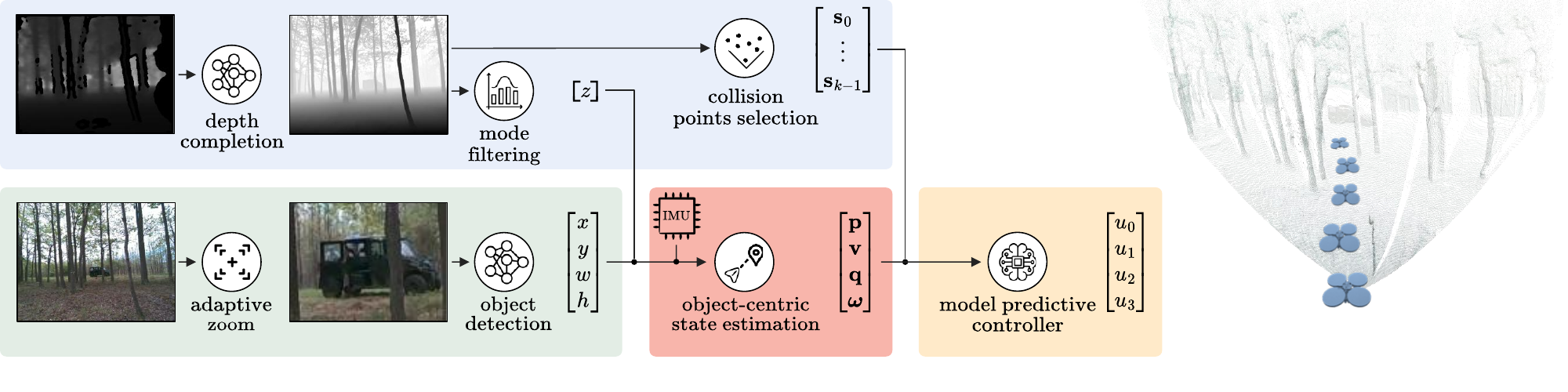}
    \caption{\textbf{System architecture overview}. The framework consists of two tightly integrated modules: perception and control. The perception stack detects the target using a lightweight object detector, estimates its depth via stereo-monocular disparity alignment, and processes obstacle information for collision avoidance. These outputs are fused with inertial data (omitted from figure for simplicity) to produce a smooth, high-frequency target-centric odometry. The control module uses NMPC, augmented with high-order CBFs, to generate safe and agile thrust and angular rate commands for tracking the target.}
    \label{fig:methodology}
\end{figure*}

\subsection{Object-Centric Visual Autonomy}
Traditional navigation frameworks rely on global localization through GPS, VIO, or SLAM, followed by waypoint-based trajectory generation~\cite{do2019high, saviolo2021autotune, kulkarni2024aerial}. While effective in structured environments, these methods often fail in GPS-denied or visually degraded settings, where feature tracking and map consistency degrade. Object-centric navigation offers a promising alternative: instead of referencing a global frame, control objectives are defined relative to specific scene objects, such as people or vehicles. This allows robots to maintain situational awareness and track targets even when global localization is unreliable or unavailable.

Several recent works have pursued this idea, though typically under restrictive assumptions that limit applicability in high-speed or unstructured environments.

An early example is \cite{lindqvist2020nonlinear}, which formulates control directly in the target’s frame of reference. While conceptually aligned with object-centric reasoning, the system relies entirely on motion capture to provide perfect state estimates for both robot and obstacle. No onboard sensing is used, and full target velocity is assumed known. This removes perception from the equation, making the setup impractical outside controlled environments.

More onboard-focused designs have emerged, such as \cite{li2023robocentric}, which uses a monocular camera and a blob detector to track a single spherical obstacle. The framework integrates perception and control onboard, but the simplicity of the sensing pipeline imposes its own constraints. Obstacle avoidance is implemented as a soft cost, tuned manually against the tracking objective, and the system assumes only one obstacle is visible at a time. The drone must be launched manually, and no mechanism is provided for adapting to clutter or dynamic changes. While a step forward, the system is difficult to generalize to real-world scenarios involving dense geometry or multiple obstacles.

Multi-agent frameworks like CoNi-MPC~\cite{zhang2023coni} take a different approach, where a UAV follows a ground robot by optimizing a relative objective. This eliminates the need for global SLAM, but again depends on full-state telemetry from the target and is tested under motion capture. The absence of onboard perception and low operational speeds (under $0.4~\si{m/s}$) further limit real-world relevance.

Later extensions add LiDAR-based obstacle sensing~\cite{zhang2025global}, improving situational awareness. However, core limitations persist. The system still relies on external tracking for the target and operates in structured, static environments at low speeds (below $1.5~\si{m/s}$). Visual perception remains absent, and the setup assumes favorable sensing conditions.

Together, these efforts reflect growing interest in object-relative control, but also reveal a gap: current systems are not designed for fast, reactive tracking in complex environments using only onboard sensing. Our framework is built to address exactly that gap. By combining real-time object detection, depth-completed state estimation, and constraint-aware planning in a fully onboard pipeline, we enable agile, target-relative control in GPS-denied, cluttered environment, without reliance on external infrastructure or prior maps.

\section{METHODOLOGY}

\subsection{System Overview}
NOVA is a fully onboard framework for vision-based aerial target tracking and collision avoidance in GPS-denied and visually degraded environments. It requires no maps, infrastructure, or prior tuning, enabling deployment in unknown and cluttered settings. The system is structured around two tightly integrated modules: perception and control (Figure~\ref{fig:methodology}).

The perception stack processes visual and inertial input to estimate the target’s relative state and detect potential collision risks in real time. These estimates feed into a NMPC that dynamically tracks the target while enforcing safety through system dynamics and real-time obstacle constraints. The entire pipeline runs onboard, supporting reactive agile flight with no external dependency.

\subsection{Perception}
The perception pipeline estimates the target's position and surrounding obstacles in real time using only onboard visual and inertial sensors. At each frame, the system detects the target, estimates its depth, and fuses this information with IMU data to maintain a full relative state. Simultaneously, it identifies potential collision points using completed depth maps. These estimates form the basis for downstream target-centric planning and control.

\subsubsection{Target Detection}
\begin{figure}[t]
    \centering
    \includegraphics[width=\linewidth, trim=0 370 550 0, clip]{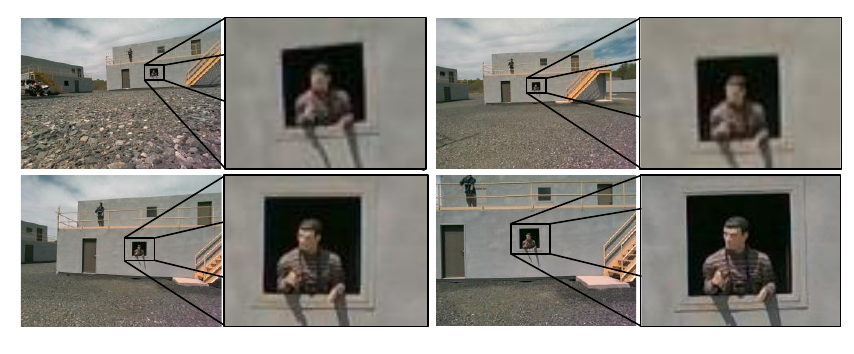}
    \caption{\textbf{Adaptive zoom strategy.} The system dynamically crops and rescales the input image to center the target while suppressing irrelevant context. This targeted zoom improves detection confidence and robustness, particularly at long range or in the presence of visually similar objects, by reducing ambiguity and limiting false positives.}
    \label{fig:zoomstrategy}
\end{figure}

A central challenge in target tracking lies in balancing detection range with computational efficiency. High-resolution inputs, such as $640 \times 480$ pixels, support target detection at distances beyond $40~\si{m}$ but exceed the processing limits of embedded platforms. In contrast, lower-resolution inputs like $320 \times 240$ pixels are more computationally efficient but constrain detection to shorter ranges, often under $10~\si{m}$. To resolve this trade-off, we introduce an adaptive zooming mechanism that dynamically crops and rescales the input image around a region of interest (Figure~\ref{fig:zoomstrategy}).

The zoom module takes the full-resolution RGB frame $\mathbf{I}^{t} \in \mathbb{R}^{H \times W \times 3}$ and a bounding box $\mathbf{b}^{t-1} = [x, y, w, h]$, where $(x, y)$ denotes the center coordinate and $(w, h)$ the width and height, from the previous detection. This box is enlarged by a factor $\alpha > 1$ to introduce a margin that accounts for motion and detection uncertainty. A crop window is then centered on the enlarged box and shaped to match the detector's aspect ratio. The zoomed-in region is extracted from $\mathbf{I}^{t}$ and resized to the detector's input resolution, yielding $\mathbf{I}^{t}_{\text{crop}}$.

If detection fails at time $t$, the crop size is progressively increased to widen the search region. Once the target is reacquired, the zoom readjusts to tightly frame the detection. This mechanism preserves long-range detection capabilities at lower resolutions while improving robustness. By narrowing the detector's receptive field to a localized region, the system avoids distractions from background clutter and allows the network to focus on the target, increasing confidence and accuracy under occlusion and motion.

We employ a lightweight, custom-trained object detector based on the YOLOv11-small architecture to enable onboard real-time detection with limited computational resources. The input to the detector is the zoomed-in crop $\mathbf{I}^{t}_{\text{crop}}$, and the output is a 2D bounding box $\mathbf{b}^t$ in crop coordinates, which is then projected back into the full-resolution frame.

The object detector is trained from scratch on the COCO dataset~\cite{lin2014microsoft} with domain-specific augmentations designed to mirror the visual effects introduced by our zooming strategy. In particular, randomized zooming and resizing simulate the pixelation and scale variations that occur during aggressive cropping, ensuring the detector remains effective even when targets appear coarse or low-resolution. Additional augmentations include pitch and roll rotations and lighting variations to improve robustness under aerial tracking conditions.

\subsubsection{Depth Estimation}
\begin{figure}[t]
    \centering
    \includegraphics[width=\linewidth, trim=0 370 550 0, clip]{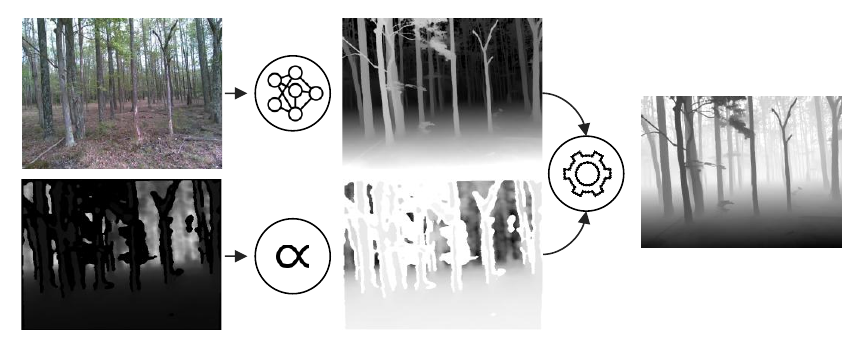}
    \caption{\textbf{Depth completion pipeline.} Monocular (top) and stereo (bottom) disparity maps are fused to produce a dense, absolute-depth estimate. Monocular predictions capture relative structure, while stereo provides scale in textured regions. A polynomial alignment module merges the two in disparity space for accurate, absolute depth.}
    \label{fig:depthcompletion}
\end{figure}

Accurate depth perception is essential for estimating the target distance as well as ensuring safe navigation. While stereo cameras provide absolute depth maps $\mathbf{D}^{t}_{\text{abs}} \in \mathbb{R}^{H \times W}$, these measurements are often sparse or noisy in areas with low texture, specular surfaces, or insufficient baseline~\cite{saviolo2024reactive}. To overcome these limitations, we employ a learning-based depth completion approach that fuses monocular and stereo cues via a disparity-domain alignment (Figure~\ref{fig:depthcompletion}).

We compute the absolute disparity from stereo depth as:
\begin{equation}
\mathbf{\Delta}^{t}_{\text{abs}} = \frac{f \cdot B}{\mathbf{D}^{t}_{\text{abs}} + \epsilon},
\end{equation}
where $f$ represents the camera's focal length, $B$ the stereo baseline, and $\epsilon$ a small constant used for numerical stability. In parallel, a monocular depth estimation neural network $\mathcal{N}_{\text{mde}}(\mathbf{I}^{t})$~\cite{yang2024depth2} predicts a relative disparity map $\mathbf{\Delta}^{t}_{\text{rel}}$.

To align scales, we fit a second-order polynomial between the two disparity maps using valid stereo pixels defined by a binary mask $\mathbf{M}^{t} \in \{0,1\}^{H \times W}$. Let $n$ be the number of valid pixels in $\mathbf{M}^{t}$. We construct a vector $\mathbf{y} \in \mathbb{R}^n$ containing the absolute disparities $\Delta^{t}_{\text{abs}}(i,j)$ at valid pixel locations. We also construct a design matrix $\mathbf{X} \in \mathbb{R}^{n \times 3}$, where each row $\mathbf{X}_k = [\Delta^{t}_{\text{rel}}(i_k,j_k)^2, \Delta^{t}_{\text{rel}}(i_k,j_k), 1]$. The vector of polynomial coefficients is denoted $\boldsymbol{\theta} = [a, b, c]^\top$.

The polynomial fit is defined as a least squares problem
\begin{equation}
\min_{\boldsymbol{\theta} \in \mathbb{R}^3} | \mathbf{X} \boldsymbol{\theta} - \mathbf{y} |^2,
\end{equation}
with the closed-form solution
\begin{equation}
\boldsymbol{\theta}^* = (\mathbf{X}^\top \mathbf{X})^{-1} \mathbf{X}^\top \mathbf{y}.
\end{equation}

The resulting polynomial parameters are applied across the full relative disparity map to generate a completed disparity estimate:
\begin{equation}
\mathbf{\Delta}^{t}_{\text{com}} = a \cdot (\mathbf{\Delta}^{t}_{\text{rel}})^2 + b \cdot \mathbf{\Delta}^{t}_{\text{rel}} + c,
\end{equation}
and converted back into a dense completed depth map:
\begin{equation}
\mathbf{D}^{t}_{\text{com}} = \frac{f \cdot B}{\mathbf{\Delta}^{t}_{\text{com}} + \epsilon}.
\end{equation}

To estimate the target's depth, a naive approach might extract a single value from the center of the detected bounding box $\mathbf{b}^t$. However, this is prone to failure under partial occlusions. For instance, the target may remain visible and correctly detected, while the center pixel falls on a background object or an occluding surface, yielding an incorrect depth estimate.

\begin{figure}[t]
    \centering
    \includegraphics[width=\linewidth, trim=0 0 0 0, clip]{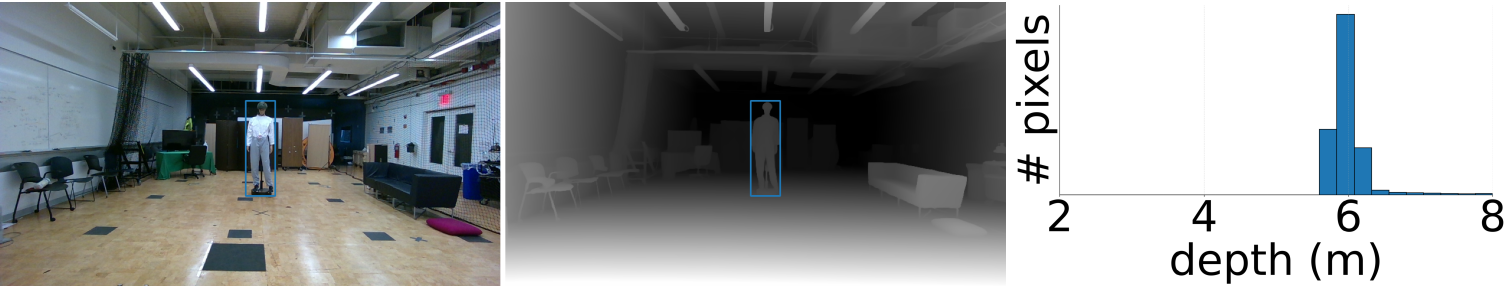}\\
    \vspace{1.05em}
    \includegraphics[width=\linewidth, trim=0 0 0 0, clip]{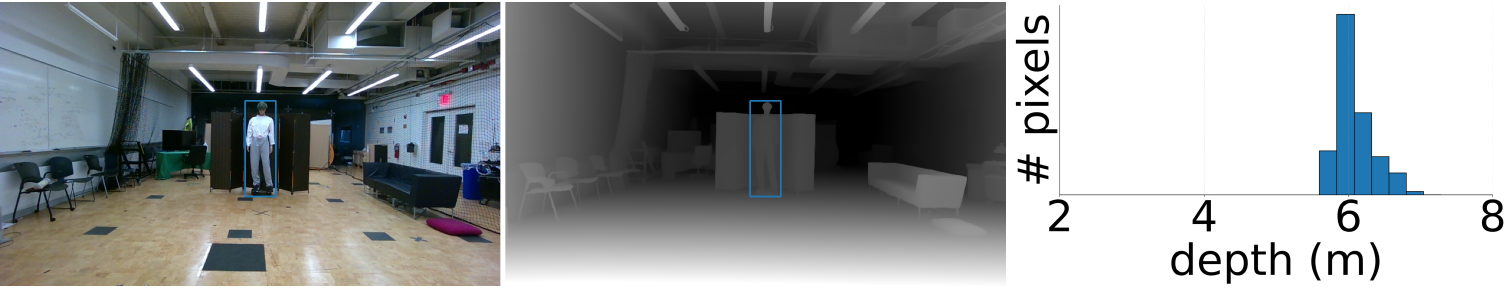}\\
    \vspace{1.05em}
    \includegraphics[width=\linewidth, trim=0 0 0 0, clip]{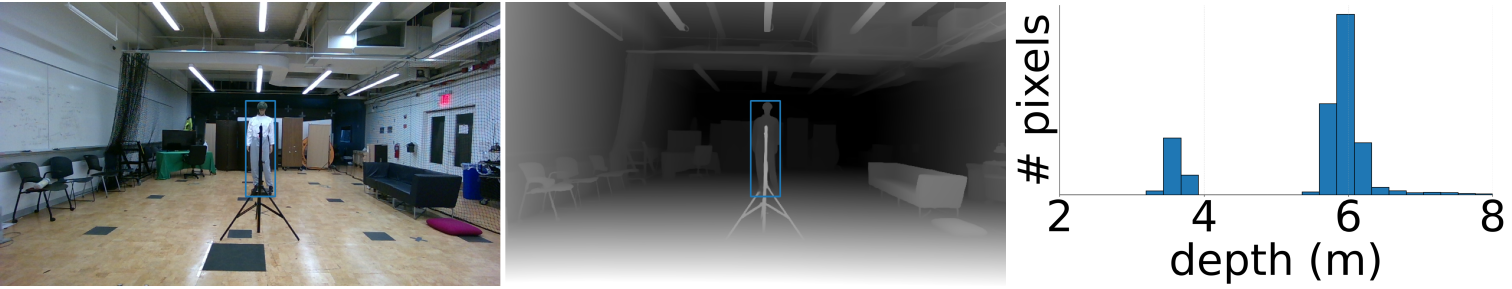}
    \caption{\textbf{Histogram-based mode filtering for target depth estimation}. Depth values inside the detected bounding box are collected and binned. The most frequent bin is selected to estimate the target distance, improving robustness to noise, occlusion, and background clutter.}
    \label{fig:depthhist}
\end{figure}

To address this, we apply histogram-based mode filtering. We extract all valid depth values within $\mathbf{b}^t$ from $\mathbf{D}^{t}_{\text{com}}$, bin them into fixed-width depth intervals, and select the most frequent bin as the representative estimate, denoted $z$. This approach leverages the observation that the dominant mode in the depth distribution corresponds to the visible target. If this was not the case, the detector would likely fail to localize the target reliably. As illustrated in Figure~\ref{fig:depthhist}, this mode-based filtering improves robustness to noise, clutter, and occlusion without requiring pixel-level semantic segmentation.

\subsubsection{Object-Centric State Estimation} \label{sec:state_estimation}
To enable smooth and reactive control, we estimate the full 6-DoF relative state of the quadrotor with respect to the target by fusing low-rate visual detections with high-rate inertial measurements. Each visual update provides the 2D center coordinate $(x, y)$ of the detected bounding box and the target depth $z$ computed via histogram-based filtering.

\begin{figure}[t]
    \centering
    \includegraphics[width=0.71\linewidth, trim=0 385 730 0, clip]{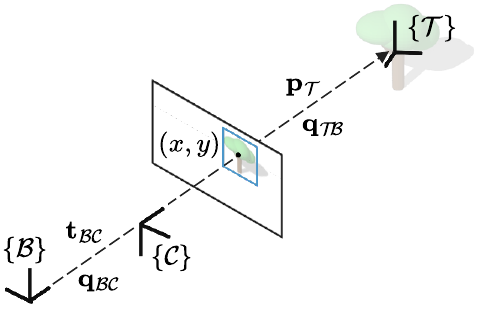}
    \caption{\textbf{Coordinate frame convention.} The object-centric frame $\mathcal{T}$ is defined at the target location, with its orientation fixed and aligned with gravity using the IMU. A 2D detection $(x, y)$ in the camera frame $\mathcal{C}$ is back-projected into 3D and transformed into the body frame $\mathcal{B}$ using known extrinsics $(\mathbf{q}_{\mathcal{BC}}, \mathbf{t}_{\mathcal{BC}})$. The resulting position is then rotated into the target frame $\mathcal{T}$ using the IMU-derived attitude $\mathbf{q}_{\mathcal{TB}}$. This defines the relative pose $\mathbf{p}_{\mathcal{T}}$, which serves as the reference for target-centric perception and control.}
    \label{fig:frameconvention}
\end{figure}

As illustrated in Figure~\ref{fig:frameconvention}, the detected 2D center coordinate is back-projected into a 3D point in the camera frame ${\mathcal{C}}$ using the known camera intrinsics $(f_x, f_y, c_x, c_y)$ obtained through camera calibration~\cite{oth2013rolling}:
\begin{equation}
\mathbf{p}_{\mathcal{C}} = \begin{bmatrix}
(x - c_x) z / f_x \\
(y - c_y) z / f_y \\
z
\end{bmatrix}.
\end{equation}

The resulting 3D point is transformed into the body frame $\mathcal{B}$ using the static extrinsics between the camera and IMU:
\begin{equation}
\mathbf{p}_{\mathcal{B}} = \mathbf{q}_{{\mathcal{BC}}} \odot \mathbf{p}_{\mathcal{C}} + \mathbf{t}_{{\mathcal{BC}}},
\end{equation}
where $\mathbf{q}_{\mathcal{BC}}$ is the camera-to-body rotation (as a quaternion), $\mathbf{t}_{\mathcal{BC}}$ the translation, and $\odot$ the quaternion-vector product.

We define a moving target-centric frame $\mathcal{T}$, centered on the target and aligned with the ground plane. 
Although this frame may translate over time, we assume it remains parallel to the ground, and we model its orientation using the IMU-derived attitude of the robot $\mathbf{q}_{\mathcal{TB}}$. The relative position of the robot in this target frame is given by:
\begin{equation}
\mathbf{p}_{\mathcal{T}} = - (\mathbf{q}_{\mathcal{TB}} \odot \mathbf{p}_{\mathcal{B}}).
\end{equation}

The resulting vector $\mathbf{p}_{\mathcal{T}}$ expresses the quadrotor's position relative to the target. This signal is tracked over time using an Unscented Kalman Filter (UKF)~\cite{loianno2016visual}.
The UKF incorporates IMU's angular velocity as process input, while visual measurements provide asynchronous updates.
The full target-centric state consists of position $\mathbf{p}_{\mathcal{T}}$, velocity $\mathbf{v}_{\mathcal{T}}$, orientation $\mathbf{q}_{\mathcal{TB}}$, and angular velocity $\boldsymbol{\omega}_{\mathcal{B}}$.

This approach eliminates the need for any environment-specific information, such as visual texture or persistent features, for state estimation. Unlike traditional VIO or SLAM systems, our method relies solely on the detection and tracking of the target to estimate the full relative state of the quadrotor. This enables robust performance even in textureless, dynamic, and unknown environments, making it especially well-suited for agile target tracking.

\subsubsection{High-Risk Collision Points Selection}
To support safe target tracking in cluttered environments, we extract a sparse set of high-risk obstacle points from the completed depth map $\mathbf{D}^{t}_{\text{com}}$ for use in downstream model predictive control. These points are selected based on a Time-To-Collision (TTC) criterion computed with respect to the robot's instantaneous velocity $\mathbf{v}_{\mathcal{T}}$.

We first generate a TTC map $\mathbf{T}^{t} \in \mathbb{R}^{H \times W}$ by projecting robot's velocity vector $\mathbf{v}^{t}_{\mathcal{C}}$, expressed in the camera frame, along the viewing ray direction of each pixel. 
These ray directions $\mathbf{R}^{t}(i, j)$ are unit-length vectors pointing from the camera center through each pixel $(i,j)$, and are computed from the intrinsic calibration parameters. 
The TTC map is then computed as:
\begin{equation}
\mathbf{T}^{t}(i, j) = \frac{\mathbf{D}^{t}_{\text{com}}(i, j)}{\|\mathbf{v}^{t}_{\mathcal{C}} \cdot \mathbf{R}^{t}(i, j)\|}.
\end{equation}

To reduce the dimensionality of $\mathbf{T}^t$, we divide it into non-overlapping grid cells $\mathcal{P}_{u,v}$ of size $P \times P$, and select the pixel in each cell with the minimum TTC:
\begin{equation}
    \mathcal{S}^{t}_{\text{ttc}} = 
    \left\{
        (i, j) \in \mathbf{T}^{t} \mid (i, j) = \argmin_{(i, j) \in \mathcal{P}_{u,v}} \mathbf{T}^{t}(i, j), \; \forall \, u, v 
    \right\}.
\end{equation}

We then apply a filtering step to retain only those points likely to intersect the quadrotor’s projected dimensions at their corresponding depths:
\begin{equation}
    \mathcal{S}^{t}_{\text{filt}} = 
    \left\{ 
        (i, j) \in \mathcal{S}^{t}_{\text{ttc}} \; \middle| \;
        \begin{aligned}
            &\left| i - c_x \right| \leq Q_x f_x / [2\mathbf{D}^{t}_{\text{com}}(i, j)], \\
            &\left| j - c_y \right| \leq Q_y f_y / [2\mathbf{D}^{t}_{\text{com}}(i, j)]
        \end{aligned}
    \right\},
\end{equation}
where $Q_x, Q_y$ represent the robot's width and height.

From this filtered set, we extract the top-$K$ most critical collision points:
\begin{equation}
    \mathcal{S}^{t}_{\text{top-}K} = 
    \argmin_{\substack{\mathcal{S'}^{t}_{\text{filt}} \subseteq \mathcal{S}^{t}_{\text{filt}}, |\mathcal{S'}^{t}_{\text{filt}}| = K}} 
    \sum_{(i, j) \in \mathcal{S'}^{t}_{\text{filt}}} 
    \mathbf{T}^{t}(i, j),
\end{equation}

The selected pixel coordinates are back-projected into 3D points in the camera frame $\mathcal{C}$:
\begin{equation}
    \mathcal{S}^{t}_{\mathcal{C}} = \left\{ 
    \left( 
    \frac{(i - c_x) z}{f_x}, \frac{(j - c_y) z}{f_y}, z 
    \right) 
    \; \middle| \; 
    \begin{aligned}
        &z = \mathbf{D}^{t}_{\text{com}}(i, j), \\
        &(i, j) \in \mathcal{S}^{t}_{\text{top-}K} 
    \end{aligned} 
    \right\}.
\end{equation}

Finally, these 3D points are transformed into the body frame $\mathcal{B}$ and then to the target-centric frame $\mathcal{T}$ using the known extrinsics:
\begin{equation}
    \mathcal{S}^{t}_{\mathcal{T}} = 
    \left\{ 
    \mathbf{q}_{\mathcal{TB}} \cdot \left(\mathbf{q}_{\mathcal{BC}} \odot \mathbf{s}_{\mathcal{C}} + \mathbf{t}_{\mathcal{BC}}\right) 
    \; \middle| \; \mathbf{s}_{\mathcal{C}} \in \mathcal{S}^{t}_{\mathcal{C}} 
    \right\},
\end{equation}

These target-centric obstacle points $\mathcal{S}^{t}_{\mathcal{T}}$ are published at each planning cycle and used to construct hard safety constraints within our NMPC formulation.

\subsection{Planning and Control}
We adopt an NMPC framework to compute optimal control commands that enable fast, smooth, and reactive flight while tracking the target and avoiding obstacles. At each time step, the NMPC solves an optimal control problem over a finite horizon, minimizing deviations from a dynamically updated target reference while ensuring dynamic feasibility and safety.

\subsubsection{Object-Centric Modeling}
We define the object-centric state of the quadrotor relative to the moving target as
\begin{equation}
    \mathbf{x} = 
    \begin{bmatrix}
        \mathbf{p}_\mathcal{T} \\[1ex]
        \mathbf{v}_\mathcal{T} \\[1ex]
        \mathbf{q}_\mathcal{TB} \\[1ex]
        \boldsymbol{\omega}_\mathcal{B} 
    \end{bmatrix} \in \mathbb{R}^{13},
\end{equation}
and the control input $\mathbf{u} \in \mathbb{R}^4$ as the vector of motor thrusts.
Thus, the quadrotor’s continuous-time dynamics evolve as
\begin{equation} \label{eq:empirical_method}
    \dot{\mathbf{x}} =
    \begin{bmatrix}
    \dot{\mathbf{p}}_\mathcal{T} \\[1ex]
    \dot{\mathbf{v}}_\mathcal{T} \\[1ex]
    \dot{\mathbf{q}}_\mathcal{TB} \\[1ex]
    \dot{\boldsymbol{\omega}_\mathcal{B}}
    \end{bmatrix} =
    \begin{bmatrix}
    \mathbf{v}_\mathcal{T} \\[1ex]
    ( \mathbf{q}_\mathcal{TB} \odot \boldsymbol{\tau} ) / m + \boldsymbol{g}_\mathcal{T} \\[1ex]
    ( \mathbf{q}_\mathcal{TB} \odot \boldsymbol{\omega}_\mathcal{B} ) / 2 \\[1ex]
    \mathbf{J}^{-1} (\boldsymbol{\mu} - \boldsymbol{\omega}_\mathcal{B} \times \mathbf{J} \boldsymbol{\omega}_\mathcal{B})
\end{bmatrix},
\end{equation}
where $m$ is the quadrotor mass, $\mathbf{g}_\mathcal{T} = \arraycolsep=2pt \begin{bmatrix} 0 & 0 & -9.81 \end{bmatrix}^\top$ is gravity in the target frame, $\mathbf{J} = diag(J_{xx}, J_{yy}, J_{zz})$ is the diagonal moment of inertia matrix, and the collective thrust $\boldsymbol{\tau}$ and torque $\boldsymbol{\mu}$ of the quadrotor are defined as
\begin{equation} \label{eq:torque}
    \boldsymbol{\tau} = k_{\tau} \sum_{i=0}^3 u_i^2, 
    \hspace{1em}
    \boldsymbol{\mu} = 
    \begin{bmatrix}
    k_{\tau} l (u_0^2 + u_1^2 - u_2^2 - u_3^2) \\[1ex]
    k_{\tau} l (-u_0^2 + u_1^2 + u_2^2 - u_3^2) \\[1ex]
    k_{\mu} (u_0^2 - u_1^2 + u_2^2 - u_3^2)
    \end{bmatrix}
    .
\end{equation}
where $k_{\tau}$ is the rotor thrust constant, $k_{\mu}$ the rotor torque constant, and $l$ the length of the quadrotor arm.

To obtain a discrete-time model, we define the dynamics function $f(\mathbf{x}, \mathbf{u}, \delta t)$ as the result of forward integration of the continuous-time dynamics over a fixed time step $\delta t$. Specifically, the next state is computed as
\begin{equation}
    \mathbf{x}^{t+1} = f(\mathbf{x}^{t}, \mathbf{u}^{t}, \delta t),
\end{equation}
where, in practice, this integration can be approximated using numerical methods such as Euler or Runge-Kutta~\cite{cartwright1992dynamics}.

This state-action-dynamics formulation is commonly used for quadrotors and provides a flexible foundation that supports sophisticated dynamic models, including aerodynamics, vibrations, motor interactions, and drag forces~\cite{saviolo2023active, saviolo2022pitcn, bauersfeld2021neurobem, crocetti2023gapt}. By adopting this formulation for visual target tracking, we can leverage the extensive prior work on dynamics modeling developed for traditional trajectory tracking tasks~\cite{saviolo2023learning}.

\subsubsection{Reference Generation}
Directly commanding the quadrotor to reach the target’s estimated position can lead to unsafe behavior, particularly under noisy observations or occlusion. To mitigate this, we implement a goal-shifting strategy that maintains a minimum horizontal safety margin, denoted $d_{\text{safe}}$, in the $xy$-plane. This creates a cylindrical exclusion zone around the target, allowing for vertical flexibility (e.g., during takeoff or hovering) while maintaining lateral separation.

The shifted reference is computed along the horizontal line-of-sight vector from the target to the current quadrotor position:
\begin{equation}
    \bar{\mathbf{p}}^{t}_{\mathcal{T}} = 
    d_{\text{safe}} \cdot 
    \frac
    {[p_{\mathcal{T}, x}, p_{\mathcal{T}, y}, 0]^\top}
    {\sqrt{p_{\mathcal{T}, x}^2 + p_{\mathcal{T}, y}^2}},
\end{equation}
ensuring that the reference remains directionally aligned in the plane while enforcing a safe lateral offset.

To avoid discontinuities during tracking and to facilitate safe takeoff behavior, this shifted goal is blended with the robot’s current position using a time-varying ramp function:
\begin{equation}
\bar{\mathbf{p}}^{t}_{\mathcal{T}} = \alpha(t) \cdot \mathbf{p}_{\mathcal{T}} + (1 - \alpha(t)) \cdot \bar{\mathbf{p}}^{t}_{\mathcal{T}},
\end{equation}
where $\alpha(t)$ is a monotonically decreasing ramp function with quadratic decay in the horizontal plane and linear decay in altitude. Initially, $\alpha(0) = 1$, anchoring the reference to the current position. As $\alpha(t) \rightarrow 0$, the reference converges to the shifted goal, ensuring smooth takeoff and tracking.

The reference trajectory passed to the NMPC is generated by replicating the same desired state over the prediction horizon of length $N$:
\begin{equation}
    \bar{\mathbf{x}}^{t}_{k} = 
    \begin{bmatrix}
        \bar{\mathbf{p}}^{t}_{\mathcal{T}} \\[1ex]
        \mathbf{0} \\[1ex]
        \bar{\mathbf{q}}^{t}_{\text{yaw}} \\[1ex]
        \mathbf{0} 
    \end{bmatrix} \quad \forall k \in [0, N),
\end{equation}
where $\bar{\mathbf{q}}^{t}_{\text{yaw}}$ is the desired yaw quaternion. The desired yaw angle $\bar{\psi}^t$ is obtained by correcting the current vehicle yaw $\psi^t$, extracted from the onboard quaternion $\mathbf{q}^{t}_{\mathcal{TB}}$, with the yaw offset that would center the target in the image:
\begin{equation}
    \bar{\psi}^{t} = \psi^{t} - \tan^{-1}\left(\frac{x - c_x}{f_x}\right).
\end{equation}

This yaw angle is then converted to a quaternion, assuming zero roll and pitch:
\begin{equation}
    \bar{\mathbf{q}}^{t}_{\text{yaw}} = 
    \begin{bmatrix}
    \cos \left( \bar{\psi}^{t}/2 \right) &
    0&
    0&
    \sin \left( \bar{\psi}^{t}/2 \right)
    \end{bmatrix}^\top
\end{equation}

The reference control inputs are simply set to zero for regularizing the control effort, $\bar{\mathbf{u}}_{k} = \mathbf{0}$ $\forall k \in [0, N)$.

\subsubsection{Cost Function}
The NMPC cost is designed to promote accurate target tracking, smooth control, and stable forward-facing flight. At each time step $k$, the stage cost is defined as:
\begin{equation}
J(\mathbf{x}, \mathbf{u}) = 
\underbrace{ \left\| \mathbf{x} - \bar{\mathbf{x}}_{k} \right\|^2_{\mathbf{Q}_x} }_{\text{State cost}} 
+ \underbrace{ \left\| \mathbf{u} - \bar{\mathbf{u}}_{k} \right\|^2_{\mathbf{Q}_u} }_{\text{Input cost}} 
+ \underbrace{  \left\| v_y^{\mathcal{B}} \right\|^2_{Q_{\text{o}}} }_{\text{Orbiting cost}},
\end{equation}
where $\mathbf{Q}_x$ and $\mathbf{Q}_u$ are positive-definite weighting matrices, and $Q_{\text{o}}$ is a scalar that penalizes lateral motion in the body frame. The term $v_y^{\mathcal{B}}$ is the lateral component of the quadrotor's velocity in its own body frame.

The first two terms are standard quadratic costs that drive the system toward the reference state while regularizing actuation effort. The third term addresses a fundamental observability limitation inherent to visual target tracking.

Specifically, when only a single target is detected, the quadrotor’s position relative to the target is observable, but its yaw orientation remains underconstrained. This creates ambiguity in the rotation about the vertical axis, allowing orbiting behaviors that preserve the perceived target location but degrade stability and control.

While previous methods~\cite{li2023robocentric} resolve this by tracking multiple targets, we argue that this assumption is overly restrictive in real-world aerial scenarios. Maintaining visibility of multiple targets while tracking one is extremely challenging in cluttered, dynamic, or degraded visual environments.

Instead, we introduce a soft regularization on $v_y^{\mathcal{B}}$, exploiting the fact that orbiting motion correlates with lateral body velocity. By penalizing this component, the NMPC naturally favors forward-facing, stable flight without additional perceptual burden. This approach is lightweight, assumption-free, and effective across diverse tracking tasks.

\subsubsection{Control Barrier Functions for Obstacle Avoidance}
To ensure safety during target tracking, we incorporate second-order CBFs into the NMPC to enforce minimum distance constraints with respect to perceived obstacles. Each obstacle point $\mathbf{s}^{t}_{\mathcal{T}} \in \mathcal{S}^{t}_{\mathcal{T}}$ is derived from depth perception and expressed in the target frame $\mathcal{T}$, consistent with object-centric control formulations proposed in recent works~\cite{saviolo2024reactive}.

For each high-risk point, we define a safety function:
\begin{equation}
h_k(\mathbf{x}^t) = \|\mathbf{s}^{t}_{\mathcal{T}} - \mathbf{p}^{t}_{\mathcal{T}}\|^2 - Q_{\text{max}}^2,
\end{equation}
where $Q_{\text{max}} = \max(Q_x, Q_y)$ is the minimum allowable distance that accounts for the robot's physical dimensions.

Safety is enforced by requiring the CBF condition:
\begin{equation}
\ddot{h}_k(\mathbf{x}^t, \mathbf{u}^t) + 2\lambda \dot{h}_k(\mathbf{x}^t) + \lambda^2 h_k(\mathbf{x}^t) \geq 0,
\end{equation}
which guarantees forward invariance of the safe set defined by $h_k(\mathbf{x}) \geq 0$. Here, $\lambda > 0$ controls the rate of enforcement.

The first derivative of the safety function is:
\begin{equation}
\dot{h}_k(\mathbf{x}^t) = 2(\mathbf{s}^{t}_{\mathcal{T}} - \mathbf{p}^{t}_{\mathcal{T}})^\top \mathbf{v}^{t}_{\mathcal{T}},
\end{equation}
and the second derivative, capturing the effect of control inputs, is given by:
\begin{equation}
\ddot{h}_k(\mathbf{x}^t, \mathbf{u}^t) = 2\|\mathbf{v}^{t}_{\mathcal{T}}\|^2 + 2(\mathbf{s}^{t}_{\mathcal{T}} - \mathbf{p}^{t}_{\mathcal{T}})^\top \dot{\mathbf{v}}_{\mathcal{T}}(\mathbf{u}),
\end{equation}
where $\dot{\mathbf{v}}_{\mathcal{T}}(\mathbf{u})$ is the acceleration of the robot in the target frame, obtained from the dynamics model $f(\mathbf{x}^t, \mathbf{u}^t, \delta t)$.

Each of these constraints is imposed at runtime for every selected obstacle point $\mathbf{s}^{t}_{\mathcal{T}}$, ensuring that the NMPC respects proximity constraints while pursuing the tracking objective. This formulation allows us to encode collision avoidance in a computationally efficient and differentiable form compatible with real-time optimization.

\subsubsection{Optimization Problem}
At each planning step, the NMPC solves a constrained optimal control problem to compute a dynamically feasible and safe control sequence over a finite horizon $N$. The objective is to minimize a cumulative cost that promotes accurate tracking, smooth control, and orbit-free motion:
\begin{equation}
\min_{\substack{
\mathbf{x}^{t}, \dots, \mathbf{x}^{t+N} \\
\mathbf{u}^{t}, \dots, \mathbf{u}^{t+N-1}}}
\sum_{j=0}^{N-1} J(\mathbf{x}^{t+j}, \mathbf{u}^{t+j}) + J(\mathbf{x}^{t+N}, \mathbf{0}),
\end{equation}
where the terminal cost is evaluated with zero control input.

The optimization is subject to the following constraints for all predictions $j \in [0, N]$ and safety constraints $k \in [0, K)$:
\begin{subequations}
    \begin{align}
        &\mathbf{x}^{t} = \hat{\mathbf{x}}^{t}, \\
        &\mathbf{x}^{t+1+j} = f(\mathbf{x}^{t+j}, \mathbf{u}^{t+j}), \\
        &\mathbf{x}_{\min} \leq \mathbf{x}^{t+j} \leq \mathbf{x}_{\max}, \\
        &\mathbf{u}_{\min} \leq \mathbf{u}^{t+j} \leq \mathbf{u}_{\max}, \\
        &\ddot{h}_k(\mathbf{x}^{t+j}, \mathbf{u}^{t+j}) + 2 \lambda \dot{h}_k(\mathbf{x}^{t+j}) +\lambda^2 h_k(\mathbf{x}^{t+j}) \geq 0.
    \end{align}
\end{subequations}

This formulation allows the NMPC to reason over future trajectories while embedding safety constraints, actuation limits, and regularization strategies into a unified and reactive control framework.

\begin{figure}[t]
    \centering
    \includegraphics[width=\linewidth, trim=0 70 0 0, clip]{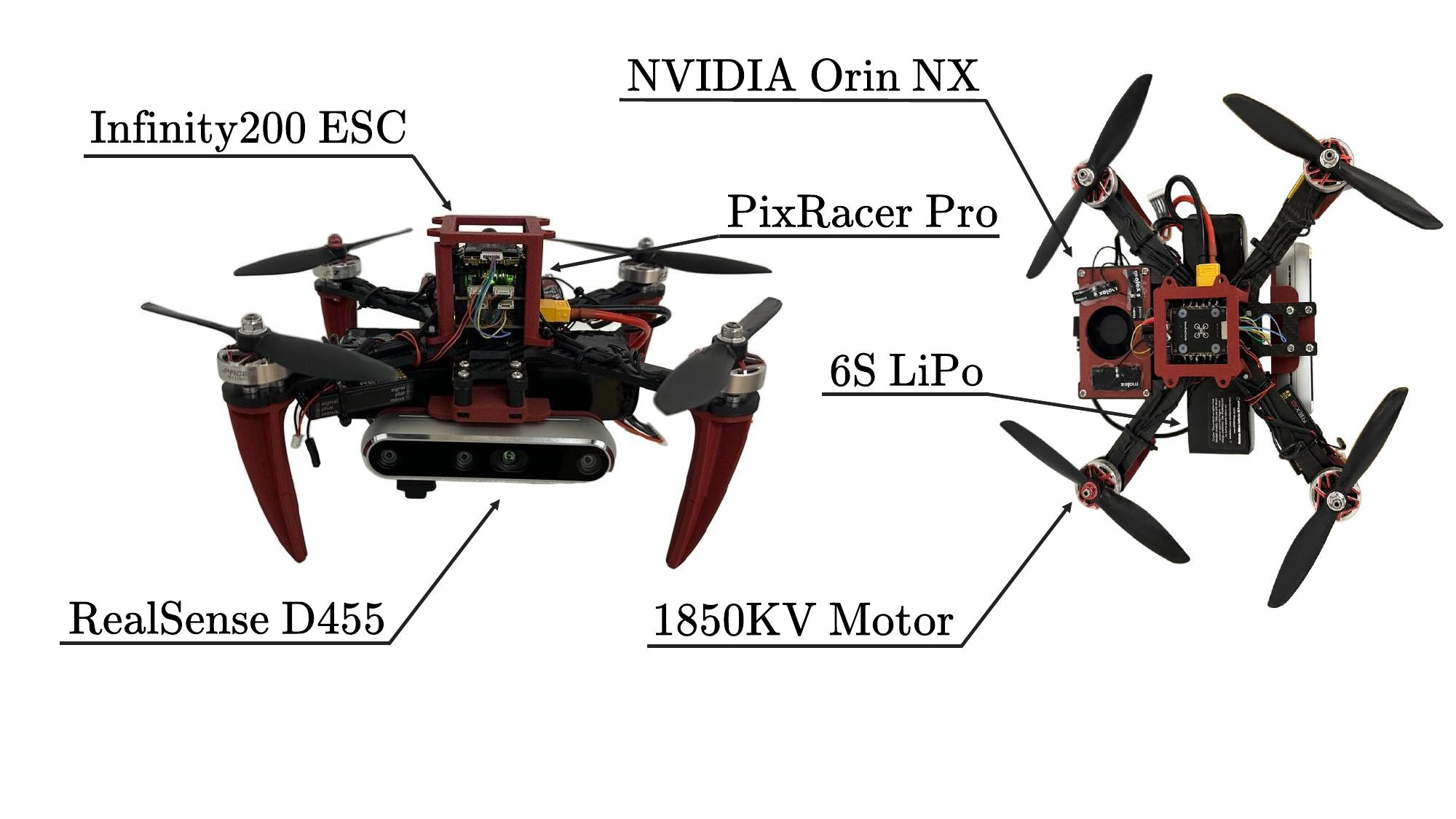}
    \caption{
    \textbf{Aerial robot platform.} Custom quadrotor used in experiments, equipped with an onboard computer, a stereo RGB-D camera, and a PX4 flight controller.
    }
    \label{fig:robot}
\end{figure}

\section{EXPERIMENTAL SETUP}
% We detail the full experimental configuration used to evaluate NOVA, including hardware specifications, sensing modalities, perception and control parameters, and training details. All system components and hyperparameters are specified to ensure reproducibility and transparency.

\subsection{Aerial System}
Our quadrotor platform, shown in Figure~\ref{fig:robot}, weighs $1.3~\si{kg}$ and has a motor span of $25~\si{cm}$, with a thrust-to-weight ratio of approximately $4$ to $1$. It is powered by a 6S LiPo battery and uses 1850KV motors paired with a NewBeeDrone Infinity200 V2 4IN1 ESC 55A to enable agile flight. 

The system integrates an Intel RealSense D455 stereo camera for visual and depth sensing and an onboard IMU. The camera is mounted front-facing, and its intrinsics are $(f_x, f_y, c_x, c_y) = (325.2, 430.9, 323.1, 246.9)$, obtained from standard calibration~\cite{oth2013rolling}. The camera-to-body transformation is defined by the static quaternion $\mathbf{q}_{\mathcal{BC}} = [-0.5, 0.5, -0.5, 0.5]^\top$ and translation $\mathbf{t}_{\mathcal{BC}} = [0.061, 0.047, -0.065]^\top~\si{m}$. RGB-D frames are captured at $60~\si{Hz}$ with a resolution of $640 \times 480$. 

Onboard processing is handled by an NVIDIA Jetson Orin NX 16GB, and low-level stabilization is managed by a PixRacer Pro flight controller.

\subsection{Target Detector}
We employ a YOLOv11-small model for object detection, trained from scratch on the COCO dataset with heavy augmentation. Images are randomly zoomed within a scale range of $[0.6, 1.4]$, rotated within $\pm15^\circ$ in pitch and roll, and augmented with synthetic motion blur ($30\%$ probability) and brightness shifts of $\pm20\%$. Crops are resized to $320 \times 256$ for inference. The model is trained using Adam with a learning rate of $2 \times 10^{-4}$, batch size $32$, weight decay $10^{-5}$, and for $300$ epochs. The inference runs at $92~\si{Hz}$ using TensorRT on the Orin NX with floating $16$ precision.

To initialize tracking, we use a one-time prompting stage that leverages a large, high-capacity object detector. Specifically, a YOLOv11x model~\cite{Jocher_Ultralytics_YOLO_2023} is run once on the full-resolution RGB image to acquire an initial bounding box for the target. This model is not trained or fine-tuned within our system. It is used solely for inference. Among all detections returned, the bounding box with the highest confidence score is selected. This bounding box seeds the zooming module for subsequent frames. After this initialization, the large YOLOv11x model is terminated to eliminate unnecessary computational overhead. From that point onward, tracking continues using only the lightweight onboard detector. Importantly, this prompting mechanism is modular. Any object detector capable of producing a coarse bounding box on the initial frame can be substituted without modifying the downstream pipeline. The sole requirement is that it provides an early spatial prior for focusing the zoomed window.

\subsection{Depth Completion}
Depth maps from the RealSense D455 (in HighAccuracy mode~\cite{grunnet2018best}) are completed using DepthAnythingV2~\cite{yang2024depth2} with a ViT-S backbone, producing per-frame estimates in $31.6~\si{ms}$. A histogram bin width of $0.15~\si{m}$ is used for mode filtering. 

\subsection{Obstacle Selection}
We divide the image into $17 \times 17$ non-overlapping cells and select $K = 10$ obstacle points with the lowest TTC. The drone's projected dimensions are $Q_x = 0.4~\si{m}$ and $Q_y = 0.2~\si{m}$. These are used in pixel-space filtering to reject low-risk points. The TTC map uses a forward-projected velocity in the camera frame and ray directions computed from calibrated intrinsics.

\subsection{State Estimation}
Target-relative state is tracked using an UKF that fuses asynchronous visual detections with high-rate inertial measurements. The filter estimates the full 6-DoF state along with accelerometer and gyroscope biases.

The UKF process noise standard deviations are defined as follows: linear acceleration noise is $(0.1, 0.1, 1.0)~\si{m/s^2}$, angular velocity noise is $(0.2, 0.2, 0.2)~\si{rad/s}$, accelerometer bias noise is $(0.1, 0.1, 0.1)~\si{m/s^2}$, and gyroscope bias noise is $(0.1, 0.1, 0.1)~\si{rad/s}$. Measurement noise standard deviations for the visual update are $0.01~\si{m}$ in $(x, y)$ and $0.001~\si{m}$ in $z$ for the position estimate, and $0.0001$ for all four quaternion components in the orientation update.

The IMU provides linear acceleration and angular velocity at $200~\si{Hz}$. The UKF is also executed at this rate and handles bias estimation online without requiring separate pre-calibration. All position and orientation estimates are computed in the target-centric frame described in Section~\ref{sec:state_estimation}.

For outdoor flights, a u-Blox Neo-M9N GPS module was fused with IMU data using an Extended Kalman Filter (EKF), yielding position updates at $100~\si{Hz}$. This global localization information was purely used for benchmarking. The proposed system operates \textbf{independently} of this information during all the flight experiments.

\begin{figure}[t]
    \centering
    \includegraphics[width=\linewidth, trim=0 200 0 40, clip]{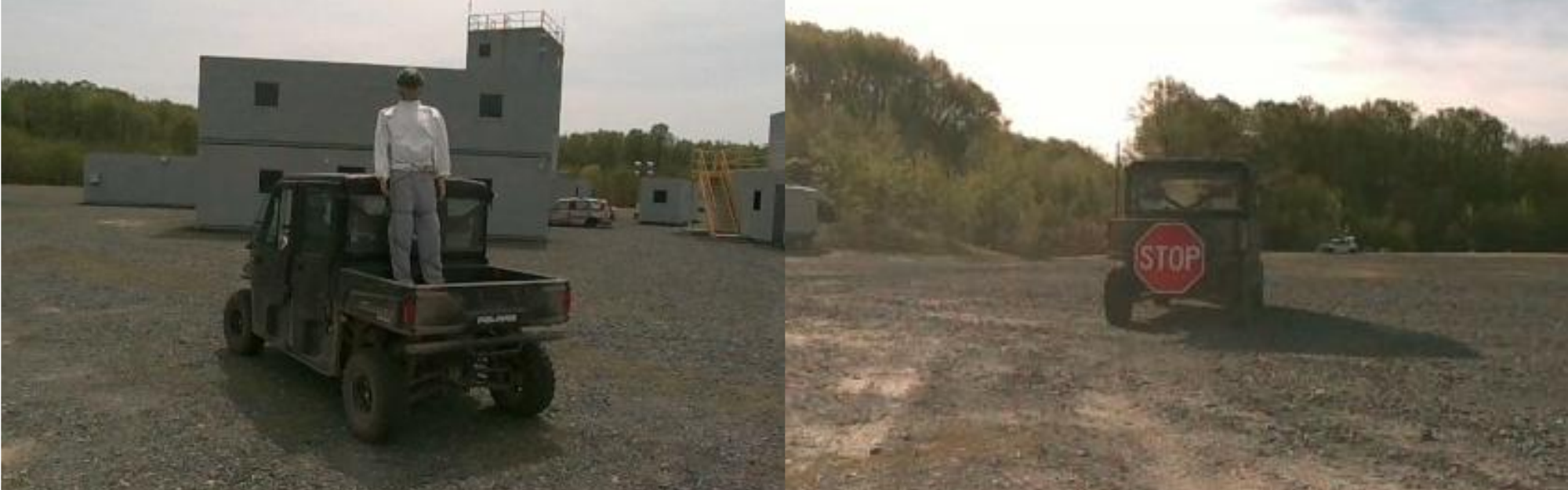}
    \caption{
    \textbf{Targets used for tracking.}
    All experiments use one of two object-mounted targets: a mannequin positioned upright on an ATV, or a stop sign mounted at the rear. The mannequin is approximately $1.8~\si{m}$ tall and wears typical clothing to simulate a human presence. The stop sign has a diameter of $0.8~\si{m}$, is red and reflective, and presents challenges due to glare and light reflections. The ATV is used to transport the targets dynamically during experiments and can reach speeds of up to $50~\si{km/h}$ across various terrains.
    }
    \label{fig:targets}
    \vspace{-1em}
\end{figure}

\begin{figure*}[tbh!]
    \centering
    \includegraphics[width=\linewidth, trim=0 35 0 20, clip]{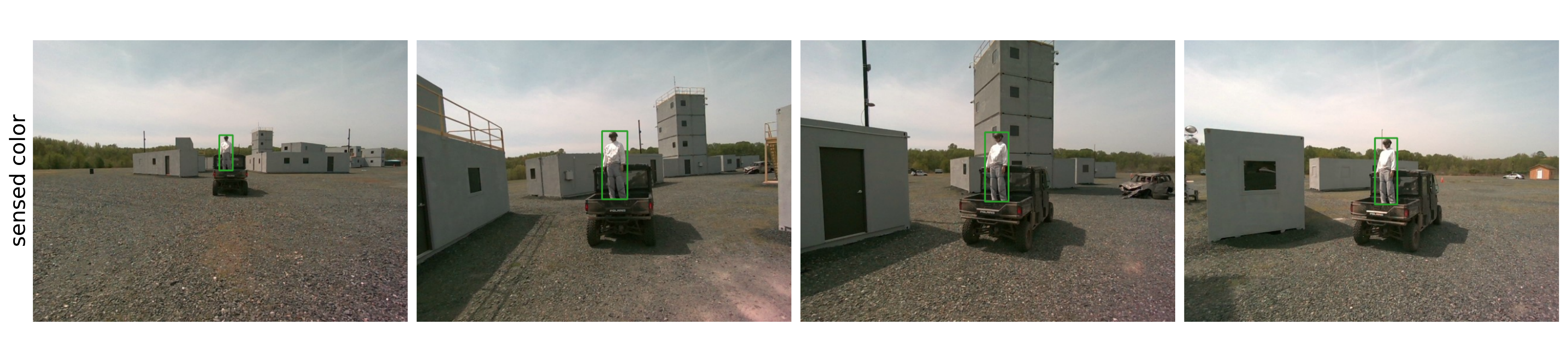}\\
    \includegraphics[width=\linewidth, trim=0 20 0 35, clip]{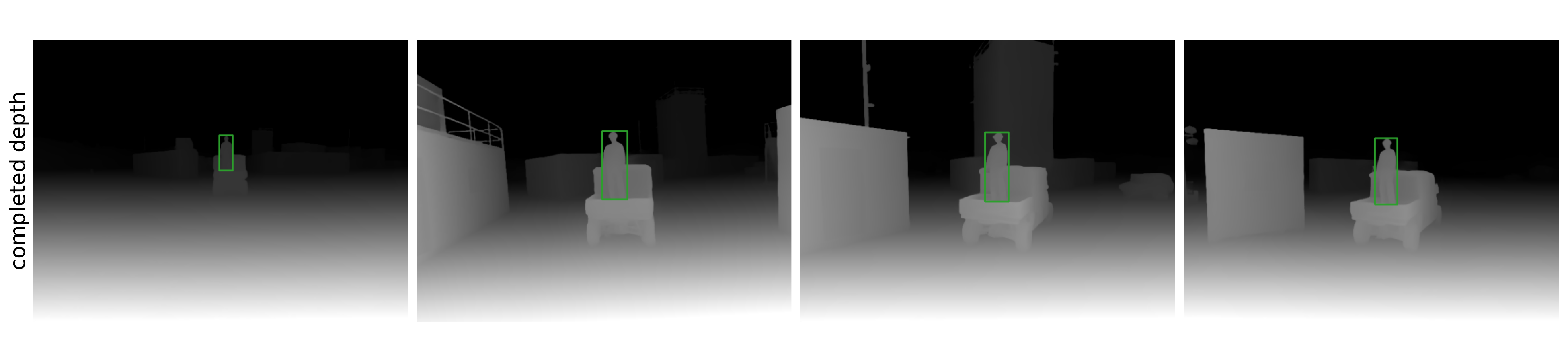}\\
    \includegraphics[width=\linewidth, trim=0 0 0 0, clip]{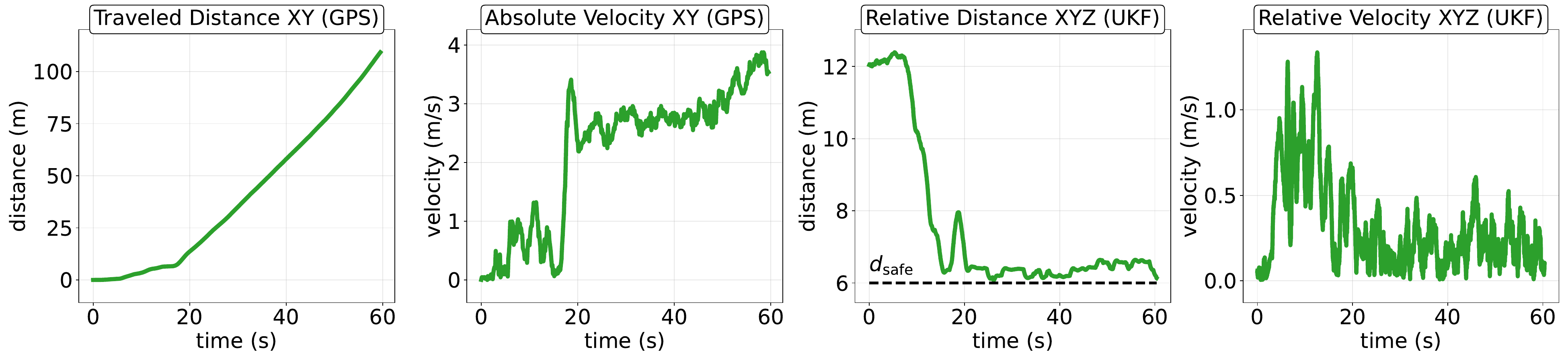}
    \caption{
    \textbf{Tracking performance in a structured urban maze.} 
    NOVA follows the target through narrow corridors and around occlusions, maintaining stable depth estimation and obstacle awareness despite minimal texture. Bottom plots show consistent velocity regulation and relative distance control, confirming safe and responsive tracking throughout the mission.
    }
    \label{fig:urban_navigation}
\end{figure*}

\subsection{Planning and Control}
We formulate an NMPC problem with a prediction horizon of $N = 10$ steps over $2~\si{s}$ and a discrete time step of $\delta t = 0.2~\si{s}$. The dynamics are integrated using a Runge–Kutta 4 method. The goal reference is shifted using a lateral safety margin of $d_{\text{safe}}$ chosen based on the experiment, and blended with a ramp function $\alpha(t)$ with quadratic horizontal decay over $1~\si{s}$. The desired yaw angle is computed to center the target in the image, and converted to a quaternion assuming zero roll and pitch.

The NMPC cost function includes a state deviation penalty $\mathbf{Q}_x = \mathrm{diag}(150, 100, 150, 15, 15, 15, 50, 15, 15, 50, 5, 5, 5)$, input cost $\mathbf{Q}_u = \mathrm{diag}(1, 1, 1, 1)$, and an orbiting penalty $Q_o = 20$ on lateral body velocity. State constraints are enforced with element-wise bounds $\mathbf{x} \in [-999, 999]^3 \times [-25, 25]^3 \times [-10, 10]^4 \times [-40, 40]^3$, and control inputs are constrained to $\mathbf{u} \in [0.05, 8.0]^4$.

Safety constraints are enforced using second-order control barrier functions with a safety radius $Q_{\text{max}} = 0.3~\si{m}$ and gain $\lambda = 2$. Up to $10$ constraints are active per solve. 

The optimization problem is solved with \texttt{acados}~\cite{acados}, using SQP with Gauss–Newton Hessian approximation and Levenberg–Marquardt regularization equal to $10^{-2}$, with a max iteration limit of $20$ and a feasibility tolerance of $10^{-4}$. 

\begin{figure*}[tbh!]
    \centering
    \includegraphics[width=\linewidth, trim=0 35 0 20, clip]{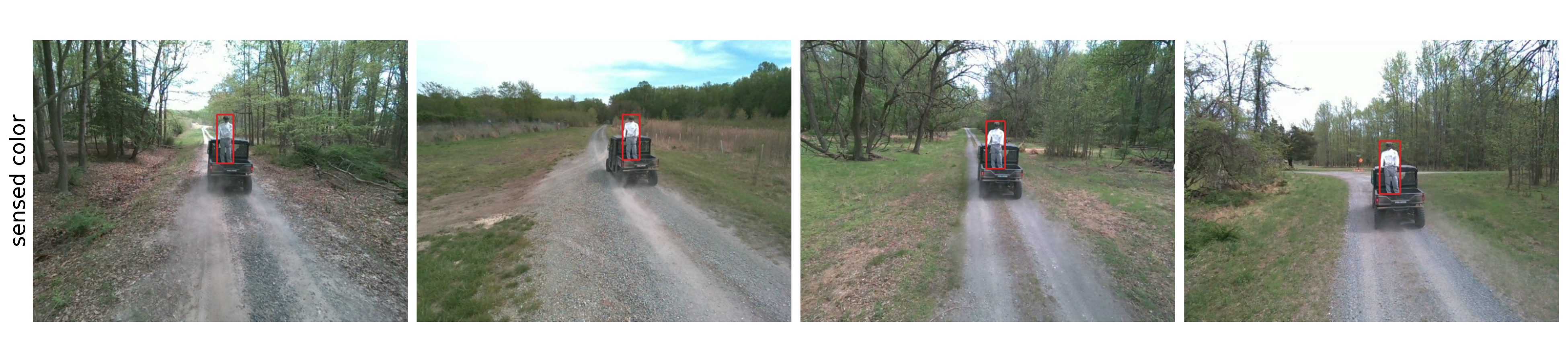}\\
    \includegraphics[width=\linewidth, trim=0 20 0 35, clip]{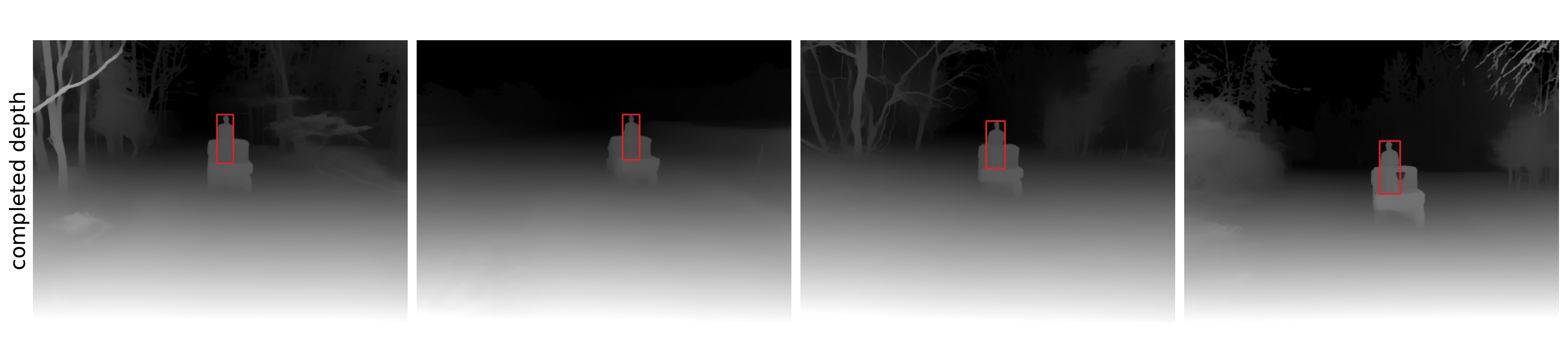}\\
    \includegraphics[width=\linewidth, trim=0 0 0 0, clip]{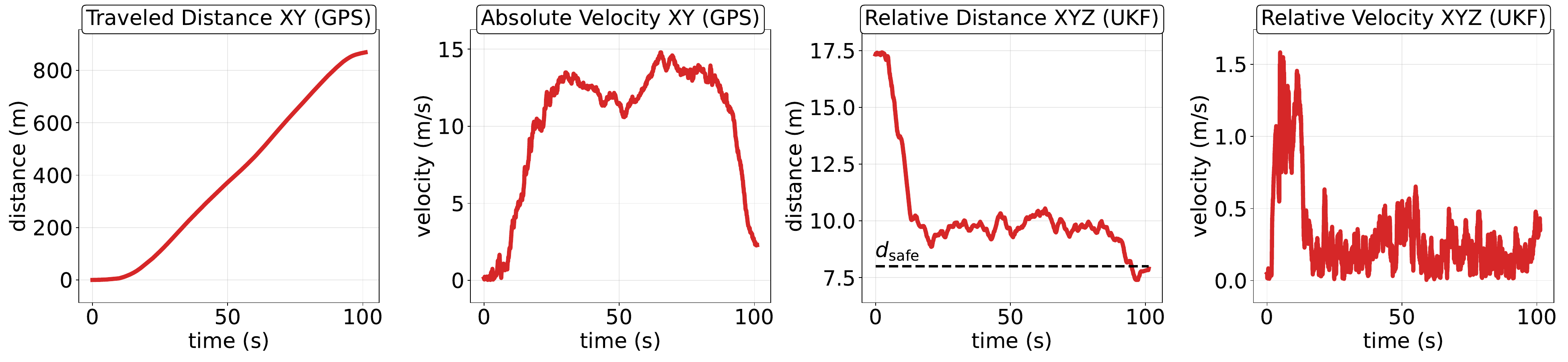}
    \caption{
    \textbf{High-speed pursuit in a rugged forest trail.}
    NOVA tracks the target over a $1~\si{km}$ unstructured path with potholes, dust, and vegetation. Despite motion blur and strong lighting variation, detection and depth estimation remain stable. Bottom plots confirm that the UAV sustains target speed while maintaining safe distance and minimal relative drift.
    }
    \label{fig:forest_trail_results}
\end{figure*}

\section{EXPERIMENTAL RESULTS}
We evaluate NOVA in a series of real-world field experiments designed to test the system’s performance under fast motion, degraded perception, and minimal prior knowledge. These tests are not isolated benchmarks, but full flight missions where the system must plan, perceive, and control entirely onboard and in real time. Our goal is to understand not only if NOVA can track a moving target, but how it behaves when the environment challenges each component of the stack.

Specifically, we focus on three central questions:
(i) Can the system remain visually coupled to fast-moving targets across large distances and terrain types?
(ii) Does the system generalize across distinct environments, including urban, forested, and indoor–outdoor transitions, without parameter retuning or prior knowledge?
(iii) Is its behavior robust and consistent across repeated trials and changes in geometry, such as altered target trajectories or forced viewpoint offsets?

To explore these questions, we deploy NOVA in environments that progressively increase in complexity. Each scenario stresses different components of the perception–control stack: visual tracking under occlusion and blur, depth estimation under lighting collapse, and planning under geometric constraint. The system runs with a fixed configuration throughout. \textit{No parameter tuning, no map registration, and no external localization is used.} All flight decisions are made from raw onboard observations in real-time onboard.

\subsection{Container Maze Navigation}
We begin by evaluating NOVA in a structured yet highly constrained setting: a container maze assembled from stacked shipping units. The layout resembles an urban canyon, with tight corridors, sharp-angle turns, and minimal escape paths. Poles, fencing, and a wrecked vehicle introduce fixed obstacles, while blind corners disrupt line of sight. The walls of the containers are flat and grey, offering little to no texture. % for stereo depth computation. 
% Compounding this, the gravel road beneath the ATV causes continuous tilt and vibration, creating transient shifts in the target’s pose that challenge both detection and estimation.

Figure~\ref{fig:urban_navigation} shows representative sequences from one flight. The RGB images capture moments when the target is partially occluded or exiting a turn. The depth maps highlight the system’s ability to extract consistent geometric information despite the lack of strong visual cues. As the UAV advances, it modulates its velocity continuously, slowing in tighter sections and accelerating when visibility improves.

The bottom plots quantify this behavior. Relative distance remains within a safe range, and the UAV’s speed adjusts tightly to match that of the ATV, even as it encounters sudden turns and obstacles. 
% Notably, this performance is achieved without any prebuilt map or localization system. NOVA relies entirely on real-time onboard sensing and prediction to maintain visual lock and safe navigation.

\subsection{Forest Trail Pursuit}
We now move into unstructured natural terrain, where sensing and control are stressed by long-range motion, fast speeds, and an unpredictable environment. The forest trail is a rugged $1~\si{km}$ path cutting through alternating segments of dense canopy and open clearings. This terrain exposes the system to rapid lighting transitions. The ground is broken and uneven, filled with potholes that jolt the ATV and introduce constant motion disturbances. Dust kicked up during traversal adds noise to the image stream, while overhanging branches and narrow passageways demand fast reactive obstacle avoidance.

The ATV reaches speeds of up to $50~\si{km/h}$, forcing the UAV to maintain high-speed flight while keeping visual lock, estimating depth, and planning safe paths through dynamic obstacles. There are no structured boundaries or predictable layouts here. NOVA must rely entirely on its onboard sensors to adapt in real time, safely plan and track the dynamic target.

As illustrated in Figure~\ref{fig:forest_trail_results}, the UAV holds tight formation throughout the run. Despite motion blur and exposure shifts, the system consistently detects the target and completes dense depth maps that preserve usable geometry. Velocity plots show near-zero relative speed, indicating that NOVA not only keeps up but tracks with high temporal fidelity. The system maintains a safe following distance with smooth speed regulation, navigating cluttered vegetation without hesitation or instability.

\begin{figure*}[t]
    \centering
    \includegraphics[width=\linewidth, trim=0 35 0 20, clip]{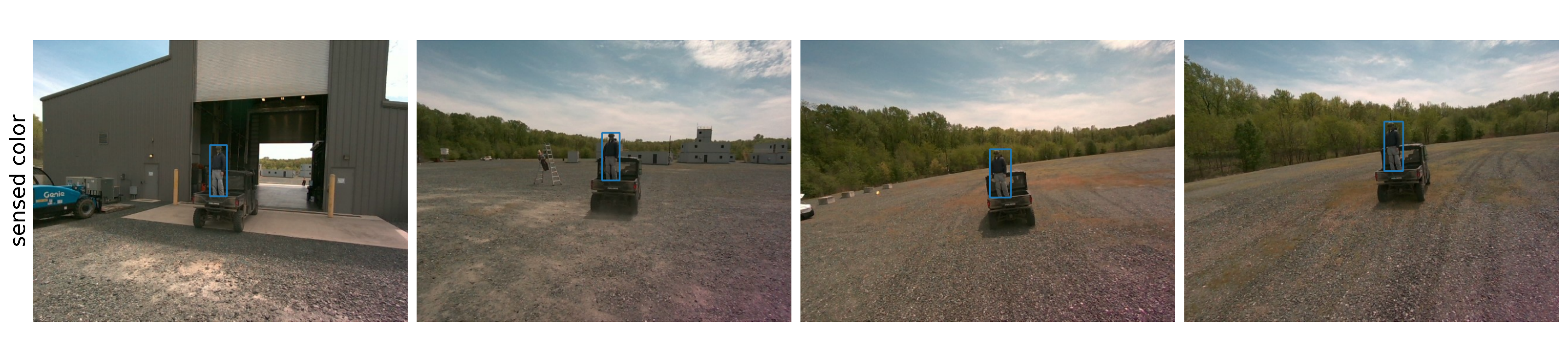}\\
    \includegraphics[width=\linewidth, trim=0 20 0 35, clip]{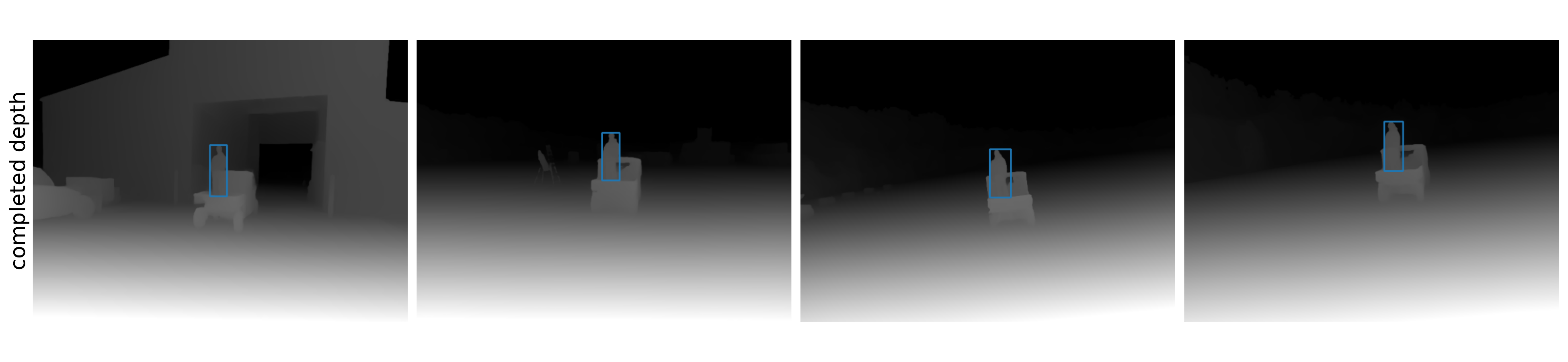}\\
    \includegraphics[width=\linewidth, trim=0 0 0 0, clip]{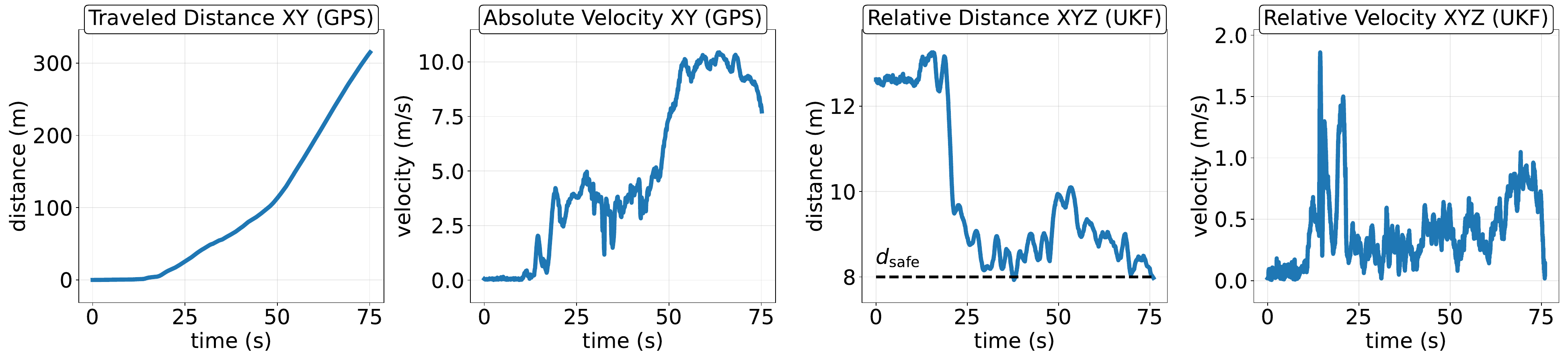}
    \caption{
    \textbf{Robust tracking across indoor–outdoor transitions.}
    The UAV follows the target from a gravel lot through a tall metallic hangar and back outside, navigating abrupt lighting changes, GPS dropout, and visual sparsity. Despite exposure shifts and structural occlusions, NOVA maintains safe distance  and continuous target lock throughout the entire mission.
    }
    \label{fig:mixed_terrain_building}
\end{figure*}

\subsection{Hangar Transition with Visual Ambiguity}
We take a step further by designing a scenario where perception is pushed to its breaking point. The mission begins in an open gravel lot under direct sunlight, transitions into a tall metallic hangar, and then returns to outdoor terrain. This path introduces a cascade of perceptual challenges: complete GPS dropout inside the hangar, abrupt lighting shifts that drive the camera from overexposure to near-total underexposure, and a cluttered interior filled with reflective surfaces, structural obstacles, and narrow doorways.

To amplify the difficulty, we place two additional mannequin targets near one of the hangar exits. These decoys are visually similar to the mannequin mounted on the ATV, requiring the system to maintain target identity precisely during a moment when visibility is degraded and the lighting is unstable. This setup explicitly tests NOVA’s robustness to visual ambiguity, occlusion, and target switching.

Such environments typically break conventional tracking pipelines. Feature-based methods struggle with low texture, stereo matching fails under reflections, and bright-to-dark transitions overwhelm standard exposure control. GPS-based localization is unavailable inside the structure, and visual odometry is easily corrupted.

Despite these conditions, NOVA maintains continuous tracking. As illustrated in Figure~\ref{fig:mixed_terrain_building}, RGB and completed depth frames remain functional even under poor lighting and low-feature geometry. When stereo cues degrade, NOVA falls back on inertial integration and histogram-filtered depth estimates to maintain a consistent relative pose. It does not require artificial markers, global maps, or external references.

The trajectory plots confirm smooth flight. Relative distance remains stable, with no abrupt control shifts or recovery maneuvers. The UAV transitions through the building without reinitialization, rejects the decoy mannequins, and maintains accurate lock on the moving target throughout the most perceptually challenging phase of the mission.

\begin{figure*}[t]
    \centering
    \includegraphics[width=\linewidth, trim=0 35 0 20, clip]{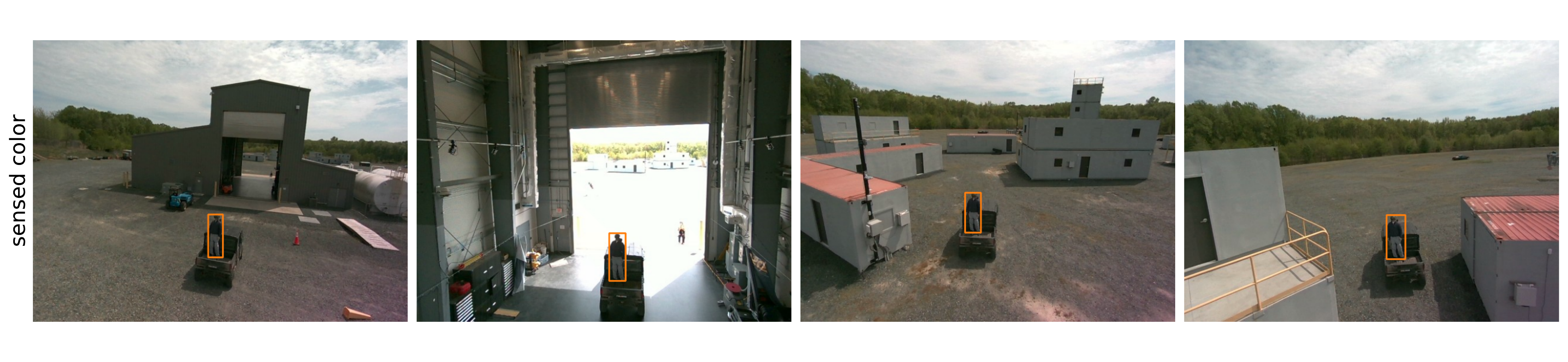}\\
    \includegraphics[width=\linewidth, trim=0 20 0 35, clip]{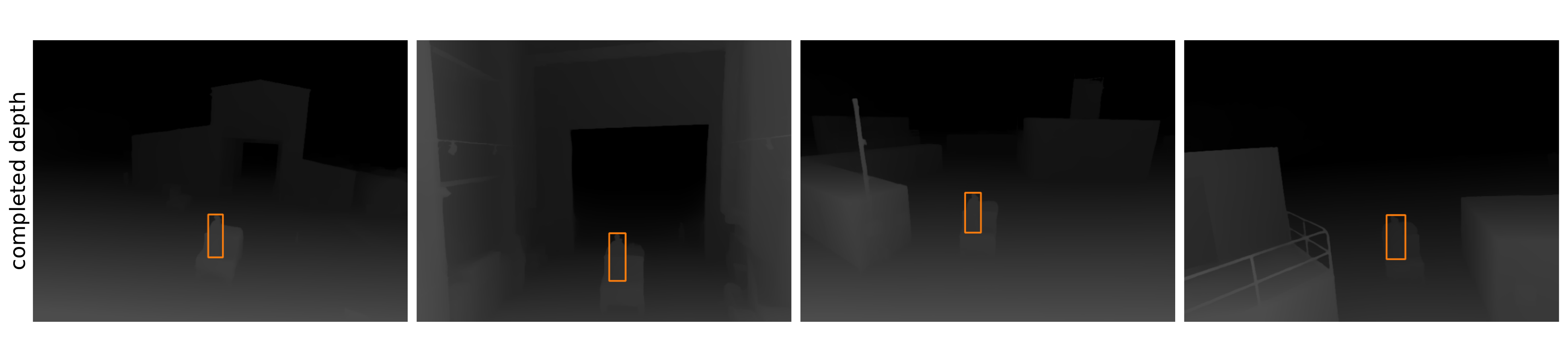}\\
    \includegraphics[width=\linewidth, trim=0 0 0 0, clip]{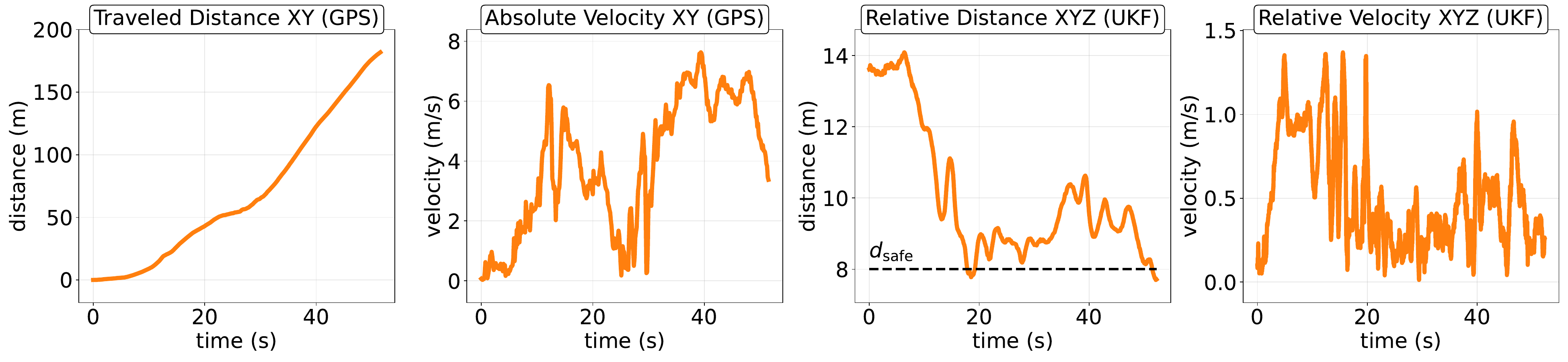}
    \caption{
    \textbf{Tracking from elevated viewpoints with forced height offset.}
    The UAV follows the target through a combined indoor–outdoor and urban maze mission while maintaining a $6~\si{m}$ vertical offset. This configuration introduces degraded stereo geometry, reduced field-of-view, and tight spatial constraints. NOVA preserves visual lock, avoids obstacles, and regulates target distance despite elevated flight and complex terrain.
    }
    \label{fig:height_offset_results}
\end{figure*}

\subsection{Elevated Tracking Across Mixed Terrain}
We take an even further step to complicate the mission by deliberately altering the spatial configuration between the UAV and the target. In this experiment, NOVA is tasked with maintaining a constant $6~\si{m}$ vertical offset from the estimated target position, resulting in a sustained flight altitude of nearly $10~\si{m}$ above ground. This constraint is evaluated end to end across a single continuous mission: from the open parking lot, through the metallic hangar, across the gravel yard, and into the corridors of the simulated container maze.

This setup introduces a compounded set of perception and control challenges. At elevated altitudes, stereo parallax is significantly reduced, degrading the quality and density of depth estimates. The target appears smaller in the frame, limiting the number of visible features and reducing detector robustness. The UAV’s field of view becomes more constrained with respect to lateral motion, making it harder to anticipate turns or sudden accelerations. Precise obstacle avoidance becomes critical, especially around overhanging structures, hangar entrances, and confined turns in the maze.

The environments themselves amplify these difficulties. The hangar imposes complete GPS dropout and rapid lighting changes. The container maze presents frequent occlusions, low-texture surfaces, and sharp-angle geometry. From a top-down viewpoint, NOVA must maintain tight target coupling while flying above clutter.

As shown in Figure~\ref{fig:height_offset_results}, NOVA completes the full mission without loss of lock or deviation from safe bounds. Despite increased vertical separation and degraded sensing geometry, it maintains coherent relative state estimates and reconstructs high-fidelity depth maps. Relative distance and velocity remain tightly regulated, and the UAV maneuvers with smooth, responsive control throughout.

This experiment confirms that NOVA generalizes not only across different environments, but also across spatial configurations that significantly alter perception and planning dynamics. No changes are made to the underlying system; all modules run as configured in prior tests, reinforcing the adaptability of the stack under shifted tracking regimes.

\begin{table*}[t]
    \centering
    \setlength{\tabcolsep}{8.8pt}
    \caption{
    \textbf{Quantitative performance across the four outdoor tracking missions.}
    We report results from representative trials in diverse environments, each introducing unique challenges in geometry, speed, sensing, and elevation. NOVA consistently maintains safe separation, stable flight, and high detection rates without environment-specific tuning. Reported metrics include UAV velocity (mean and peak), linear acceleration, pitch angle, mean and minimum distance to the target compared to the safety threshold $ d_{\text{safe}} $, and overall detection rate.
    }
    \label{tab:outdoor_tracking_results}
    \begin{tabular}{lcccccccccc}
    \toprule\toprule
    \multirow{2}{*}{\textbf{Scenario}} & \multicolumn{2}{c}{\textbf{GPS Velocity (km/h)}} & \multicolumn{2}{c}{\textbf{Control Effort (m/s²)}} & \multicolumn{2}{c}{\textbf{Pitch (°)}} & \multicolumn{3}{c}{\textbf{Rel. Distance (m)}} & \multirow{2}{*}{\textbf{Detections (\%)}} \\
    \cmidrule(lr){2-3} \cmidrule(lr){4-5} \cmidrule(lr){6-7} \cmidrule(lr){8-10}
    & \textbf{Mean} & \textbf{Max} & \textbf{Mean} & \textbf{Max} & \textbf{Mean} & \textbf{Max} & \textbf{Mean} & \textbf{Min} & $d_\text{safe}$ & \\
    \midrule
    Urban Maze          & 5.7 & 14.0 & 10.2 & 32.4 & 2.3 & 15.0 & 8.2 & 6.2 & 6.0 & 98.6 \\
    Forest Trail        & 35.0 & 53.3 & 10.2 & 13.9 & 10.1 & 19.6 & 10.0 & 8.4 & 8.0 & 94.5 \\
    Building Transition & 17.6 & 37.6 & 10.3 & 22.6 & 5.3 & 16.6 & 9.9 & 8.9 & 8.0 & 96.2 \\
    Elevated Tracking   & 14.5 & 27.5 & 10.3 & 22.8 & 7.7 & 28.8 & 9.4 & 8.8 & 8.0 & 93.1 \\
    \bottomrule\bottomrule
    \end{tabular}
\end{table*}

\begin{table}[t]
    \centering
    \caption{
    \textbf{Tracking performance consistency across multiple indoor–outdoor trials.}
    Metrics include mean and minimum relative distance, user-defined safety threshold, and relative velocity statistics. NOVA maintains stable performance across all trials without tuning or adaptation.
    }
    \vspace{0.5em}
    \begin{tabular}{c c cccc c}
        \toprule\toprule
        \multirow{2}{*}{\textbf{Trial}} & \multirow{2}{*}{\textbf{Direction}} & \multicolumn{3}{c}{\textbf{Relative Distance (m)}} & \multicolumn{2}{c}{\textbf{Rel. Velocity (m/s)}} \\
        \cmidrule(lr){3-5} \cmidrule(lr){6-7}
        & & \textbf{Mean} & \textbf{Min} & $d_{\text{safe}}$ & \textbf{Mean} & \textbf{Max} \\
        \midrule
        1 & Forward & 9.9 & 8.9 & 8.0 & 0.1 & 0.5 \\
        2 & Forward & 9.5 & 8.7 & 8.0 & 0.2 & 0.4 \\
        3 & Reverse & 9.1 & 8.9 & 8.0 & 0.1 & 0.5 \\
        4 & Reverse & 9.3 & 8.6 & 8.0 & 0.2 & 0.5 \\
        \bottomrule\bottomrule
    \end{tabular}
    \label{tab:repeatability}
\end{table}

\subsection{Measured Performance across Tested Environments}
To complement the qualitative results, we report quantitative metrics for the previous four representative experiments. Table~\ref{tab:outdoor_tracking_results} summarizes performance across these trials, covering GPS-based velocity, control effort, pitch dynamics, target distance regulation, and object detection consistency.

In the Urban Maze, the UAV operates within narrow corridors and sharp turns, demanding precise obstacle avoidance and frequent reacceleration. NOVA sustains high detection reliability ($98.6\%$) while maintaining an average separation of $8.2~\si{m}$ from the target, with a minimum of $6.2~\si{m}$, just above the $d_{\text{safe}} = 6.0~\si{m}$ threshold. The high peak acceleration of $32.4~\si{m/s^2}$ and max pitch of $15.0^\circ$ reflect sharp control inputs required to navigate tight geometry while maintaining visual lock.

The Forest Trail scenario emphasizes sustained high-speed tracking over unstructured terrain. The target exceeds $53~\si{km/h}$, and the UAV maintains a mean relative distance of $10.0~\si{m}$, dipping to $8.4~\si{m}$ at its closest—well within the $d_{\text{safe}} = 8.0~\si{m}$ constraint. Control demands are moderate, with limited acceleration spikes and a max pitch of $19.6^\circ$, indicating smooth flight despite blur, shadows, and environmental noise. NOVA’s object detector remains effective under these fast, noisy conditions, with $94.5\%$ detection accuracy.

In the Building Transition trial, the system experiences full GPS dropout and sharp illumination shifts when passing through a metallic hangar. NOVA responds with increased control activity, reaching $22.6~\si{m/s^2}$ peak acceleration and maintaining an average pitch of $5.3^\circ$. The UAV holds an average target distance of $9.9~\si{m}$, with a minimum of $8.9~\si{m}$, both within the $d_{\text{safe}} = 8.0~\si{m}$ threshold, while sustaining visual target lock at $96.2\%$.

The Elevated Tracking experiment introduces an artificial vertical offset, forcing the UAV to maintain a higher flight altitude. At these distances, stereo depth estimation becomes less reliable and target scale diminishes in the image. NOVA compensates with stronger dynamic control, producing a maximum pitch of $28.8^\circ$ and acceleration up to $22.8~\si{m/s^2}$. The UAV sustains a mean distance of $9.4~\si{m}$ and minimum of $8.8~\si{m}$, again within the $d_{\text{safe}} = 8.0~\si{m}$ limit, and maintains $93.1\%$ detection despite reduced resolution and more cluttered backgrounds.

In all cases, NOVA avoids collisions, keeps the target visibility, and respects user-defined separation thresholds. These results highlight the system’s robustness across varied real-world conditions.

\begin{figure}[t]
    \centering
    \includegraphics[width=\linewidth, trim=0 100 400 0, clip]{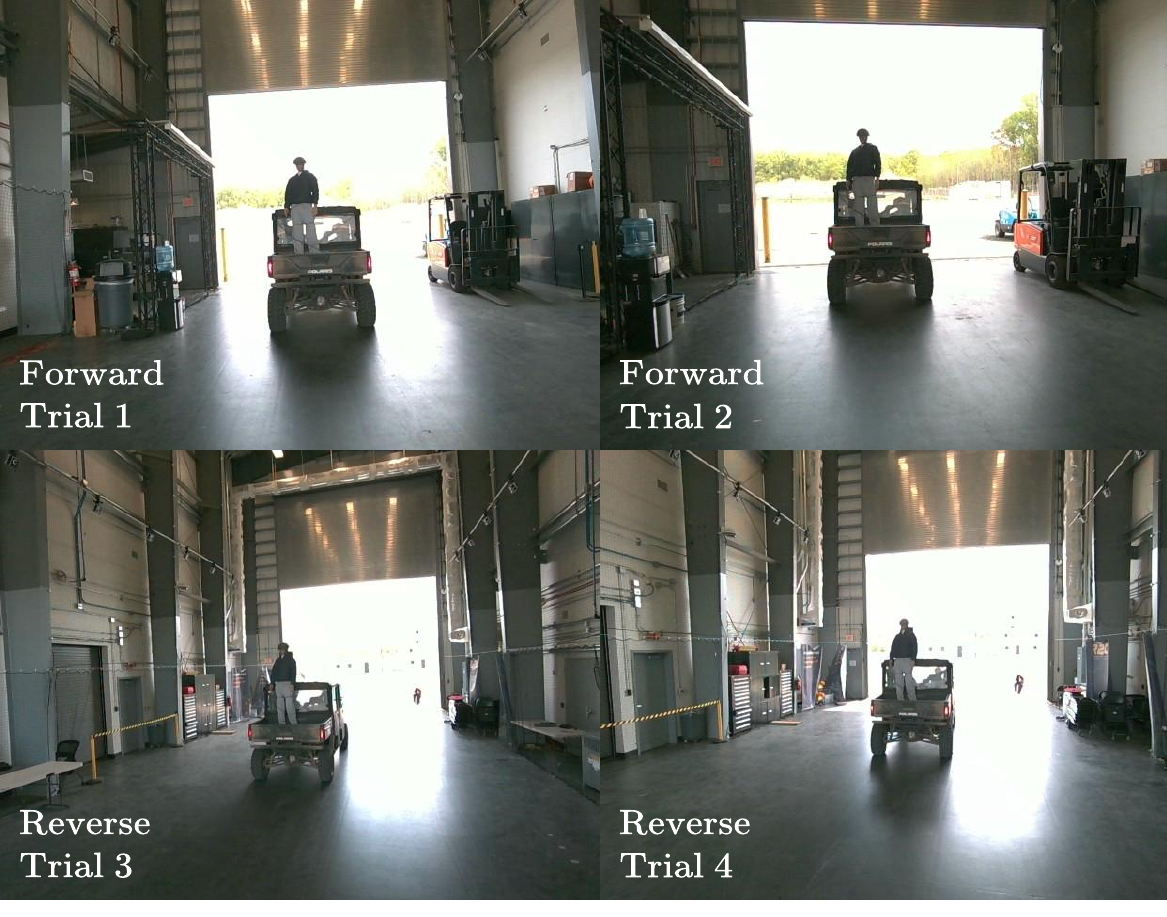}
    \caption{
    \textbf{Robust target tracking across repeated trials.} 
    Sample RGB frames from four separate indoor–outdoor transition experiments. NOVA maintains consistent target lock despite variations in lighting, entry angle, and background.
    }
    \label{fig:repeatability}
\end{figure}

\begin{figure*}[t]
    \centering
    \includegraphics[width=\textwidth, trim=0 0 0 0, clip]{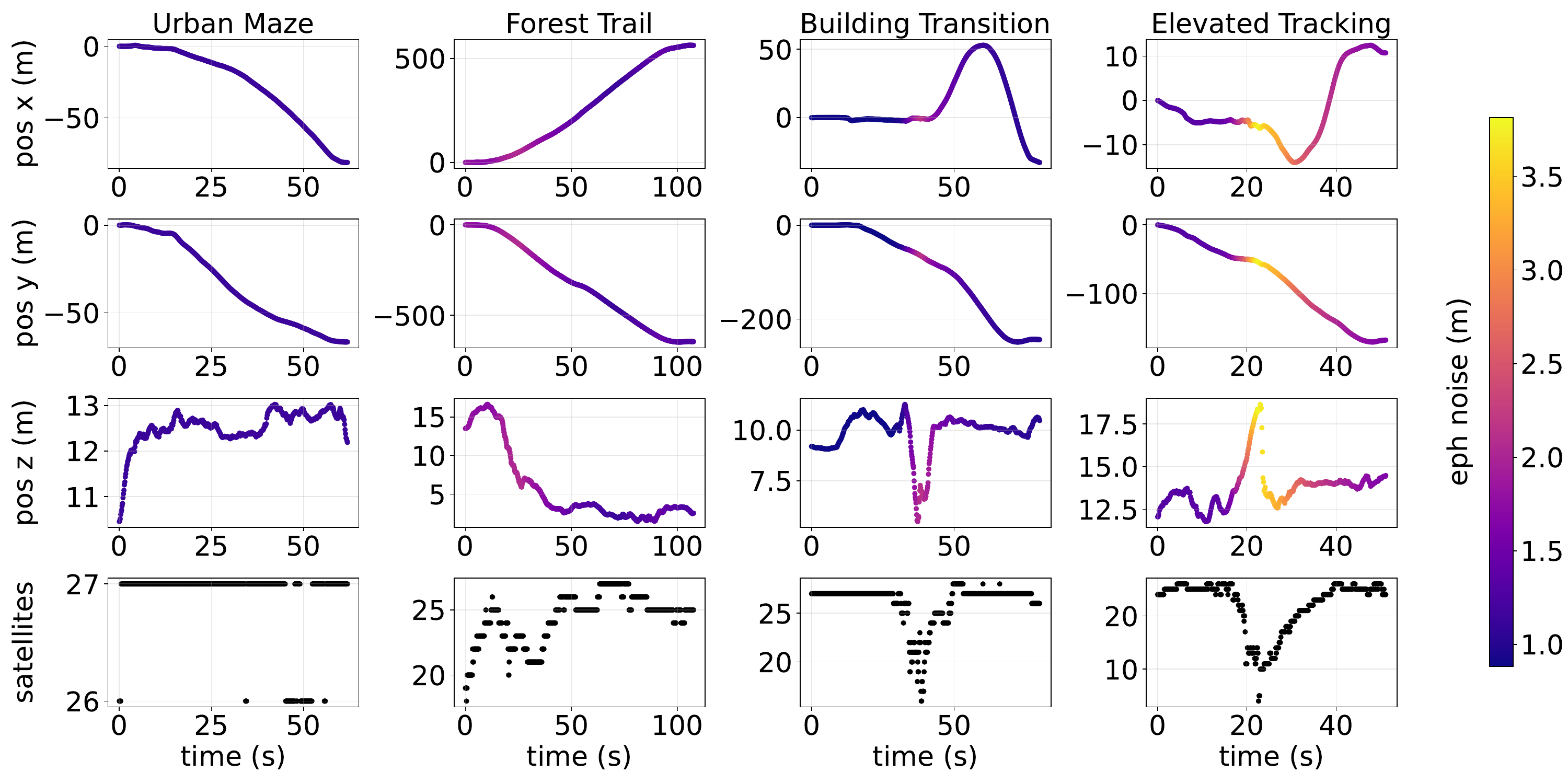}
    \caption{
    \textbf{GPS signal degradation across outdoor tracking scenarios.}
    The top rows show fused GPS+IMU global position traces for each of the four outdoor missions, color-coded by estimated horizontal positional error (eph). Lighter colors indicate higher uncertainty. The bottom plot shows the number of GPS satellites tracked over time. In the Urban Maze and Forest Trail scenarios, signal quality remains high, with low positional error and stable satellite lock. In contrast, the Building Transition and Elevated Tracking missions exhibit significant degradation, including extended satellite dropout and increased eph. This effect is especially pronounced during indoor segments and when the UAV flies near structural ceilings, highlighting the limitations of GPS-based localization in partially enclosed or cluttered environments.
    }
    \label{fig:outdoor_xyz_noise}
\end{figure*}

\subsection{Repeatability and Robustness}
Robust operation in the real world requires more than handling isolated challenges. A tracking system must perform consistently across repeated trials, despite minor variations in conditions. To evaluate this, we repeat the full indoor–outdoor transition experiment four times under nominally identical setups. Two trials proceed in the forward direction, while two others are conducted in reverse, introducing mirrored geometry, altered lighting transitions, and different entry angles.

Despite these differences, NOVA maintains stable and consistent behavior. Figure~\ref{fig:repeatability} presents sample onboard images from each run, illustrating reliable visual tracking under changes in viewpoint, shadowing, and background. No manual intervention is required between trials.

Quantitative results in Table~\ref{tab:repeatability} confirm this consistency. Across all runs, the UAV maintains safe separation well below the user-defined threshold of $d_{\text{safe}} = 8.0~\si{m}$, while keeping relative velocity and average spacing tightly bounded. Importantly, the reversed-direction flights perform on par with the forward cases, indicating that NOVA does not depend on environment-specific priors.

\subsection{GPS Signal Quality and Its Limitations}
Although NOVA operates without GPS during flight, we use fused GPS+IMU estimates post hoc to visualize global trajectories and assess how environmental geometry affects localization quality. This provides a comparative baseline for understanding where GPS-based methods remain reliable and where they degrade.

Figure~\ref{fig:outdoor_xyz_noise} summarizes GPS performance across all four representative scenarios. The top rows show the global position trace, color-coded by estimated horizontal error (eph). The bottom plot reports the number of visible satellites over time. Together, these metrics reflect how surrounding structures influence satellite visibility and positioning accuracy.

In the Urban Maze and Forest Trail scenarios, satellite visibility remains high, typically above 20 satellites, and the estimated horizontal error stays below $1.0~\si{m}$. These environments, although partially obstructed, allow for relatively consistent signal reception. The GPS data in these trials remains smooth and usable throughout the mission.

In contrast, the Building Transition and Elevated Offset experiments exhibit significant signal degradation. During the Building Transition mission, satellite count drops sharply as the UAV enters the hangar, with corresponding increases in estimated error, often exceeding $3.5~\si{m}$. These effects are due to occlusion, multipath reflections, and direct signal loss.

The Elevated Tracking scenario presents even more severe GPS degradation. Although the trajectory includes both indoor and outdoor segments, the required vertical offset places the UAV closer to the ceiling during the hangar phase. This reduces the already limited visibility to the sky and exacerbates signal dropout. In several segments, the number of visible satellites falls to near zero, and position estimates become highly uncertain.

These observations highlight the limits of GPS-based localization in built environments. Even in outdoor settings, partial occlusion or reflective surfaces can compromise satellite visibility and introduce substantial pose uncertainty. NOVA’s design avoids these failure modes by relying exclusively on onboard visual and inertial sensing, which remains operational across all tested scenarios regardless of external infrastructure availability.

\begin{figure*}[t]
    \centering
    \includegraphics[width=0.325\linewidth]{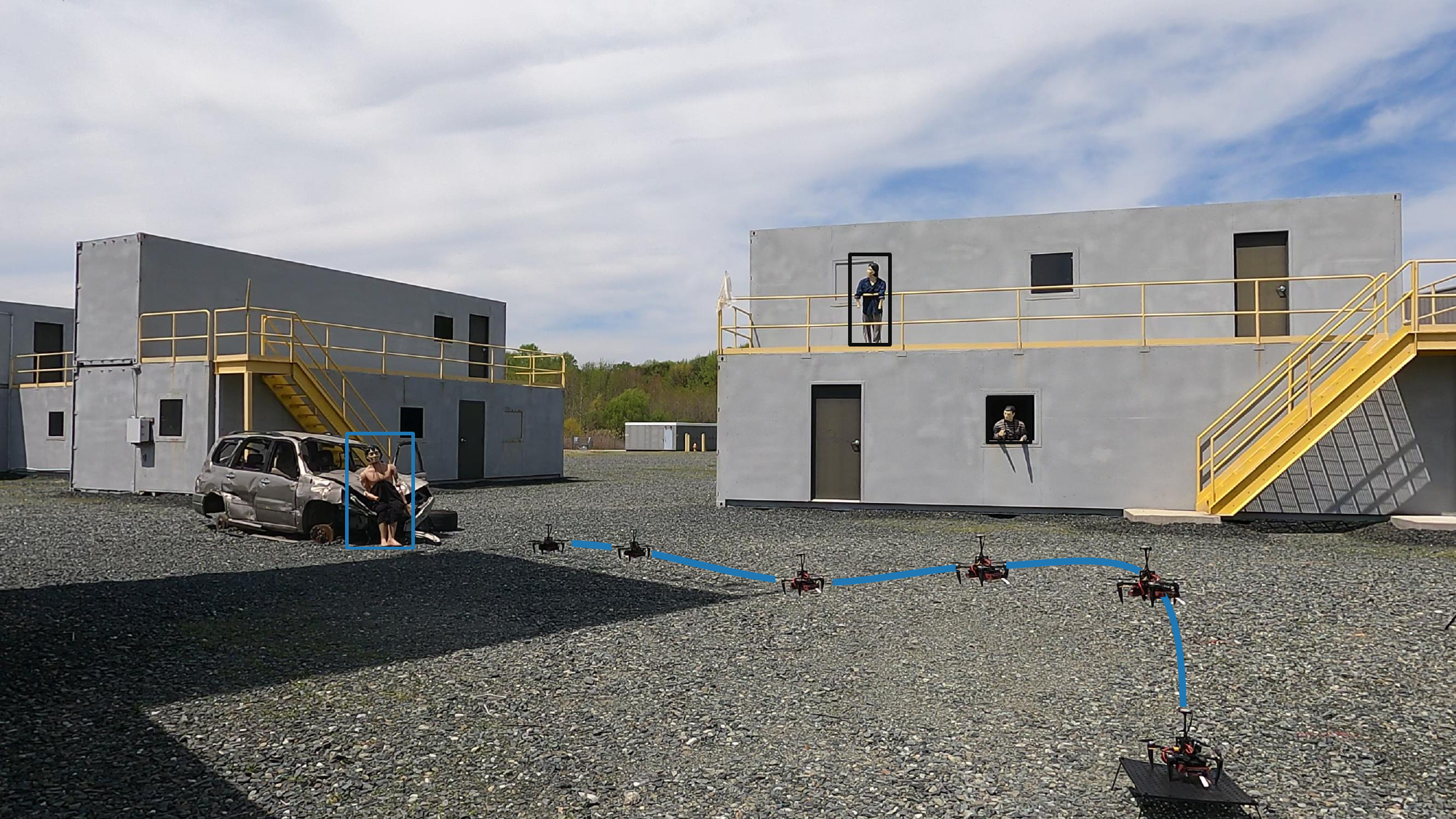}
    \hfill
    \includegraphics[width=0.325\linewidth]{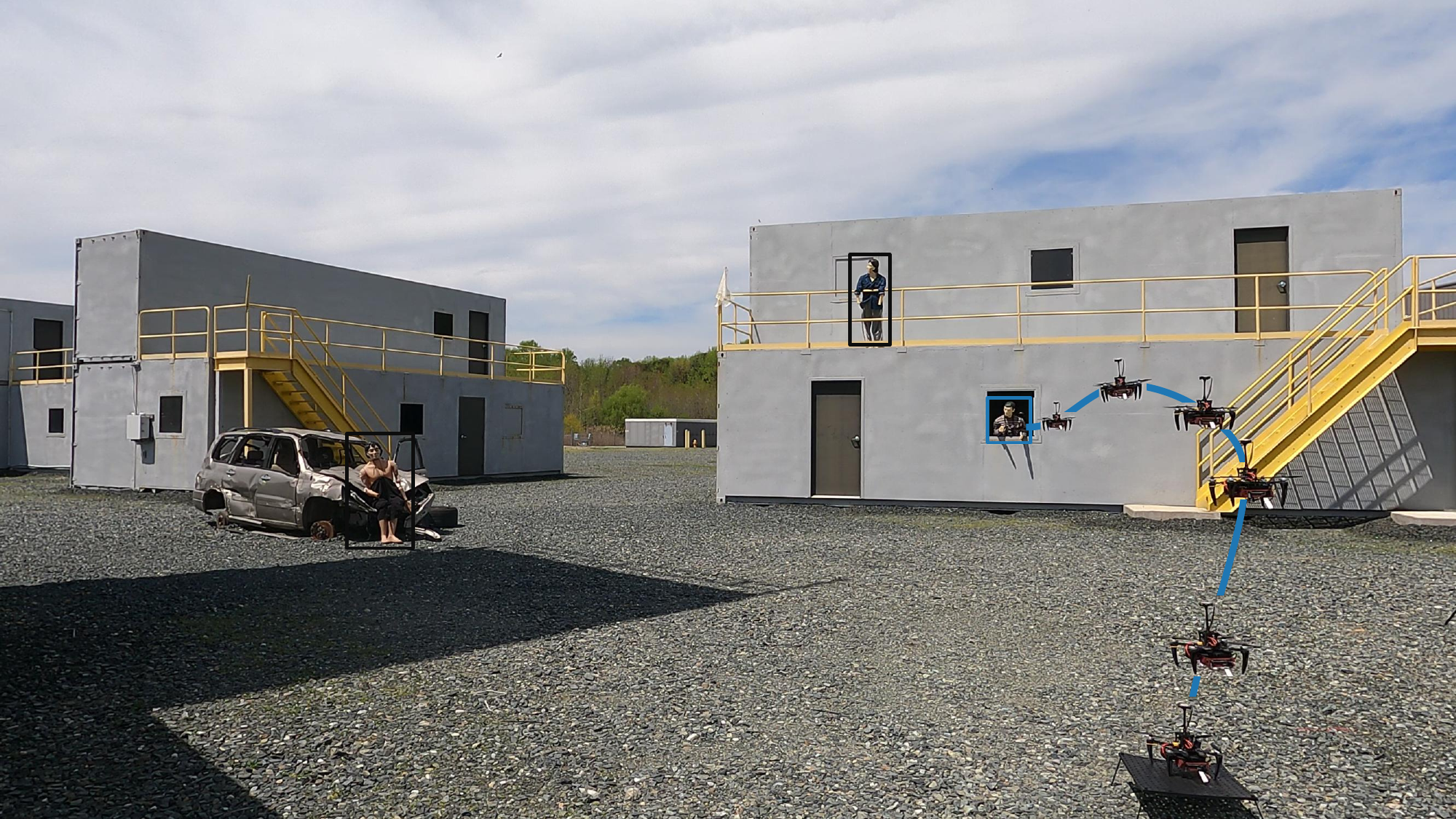}
    \hfill
    \includegraphics[width=0.325\linewidth]{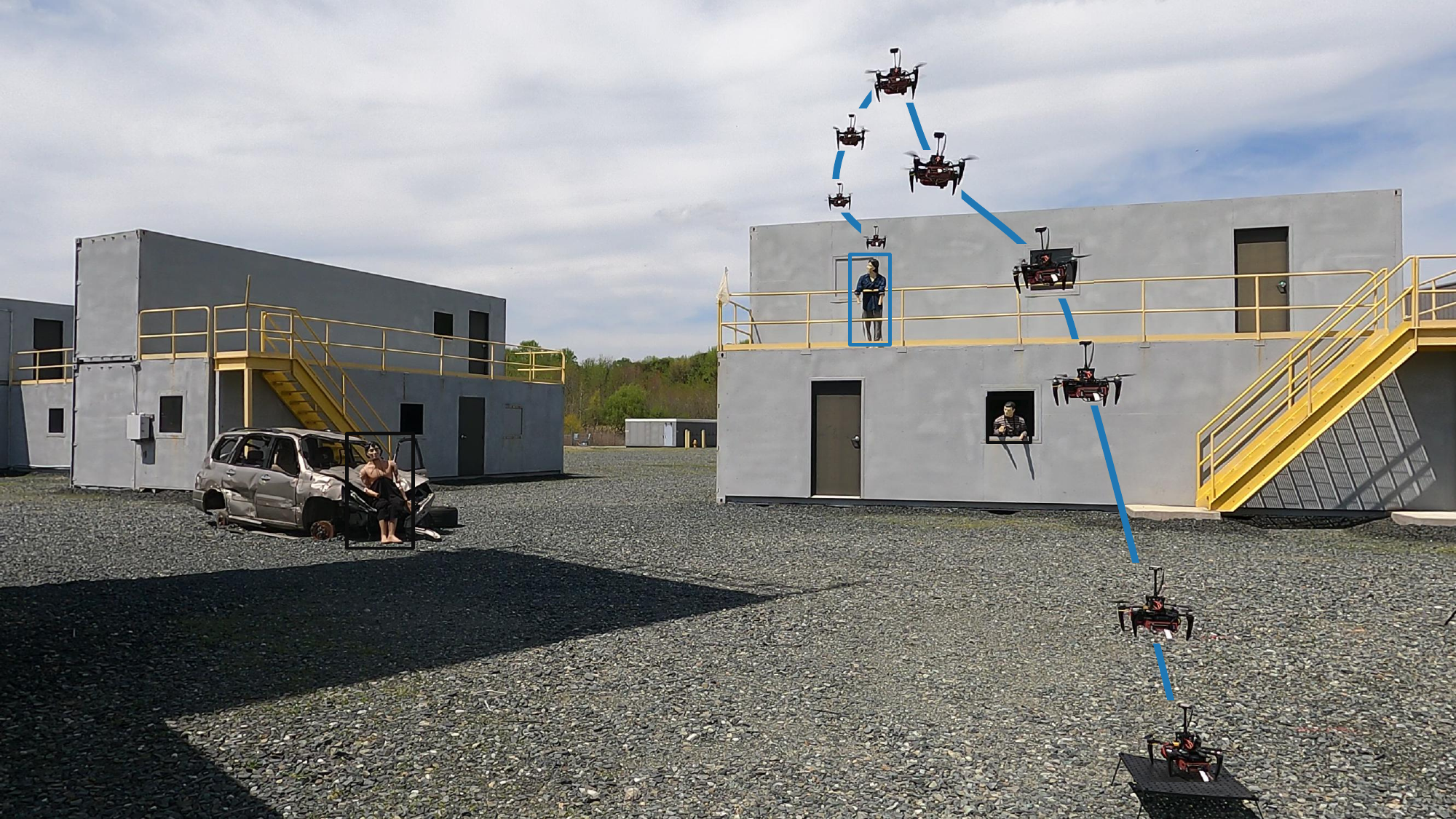}
    \caption{
    \textbf{Impact of Adaptive Zoom in Multi-Target Scenes.}
    In cluttered scenes with multiple visually similar targets (black bounding boxes), tracking can become unstable when relying on full-frame detection, often resulting in identity switches or drift. The adaptive zoom module addresses this by cropping tightly around the prompted target, suppressing distractors and maintaining detection focus. Shown here are drone trajectories (blue) for three separate trials, each prompted to track a different mannequin. NOVA successfully adheres to the correct target in all cases.
    }
    \label{fig:zoom_overlay}
\end{figure*}

\begin{figure}[t]
    \centering
    \includegraphics[width=\linewidth]{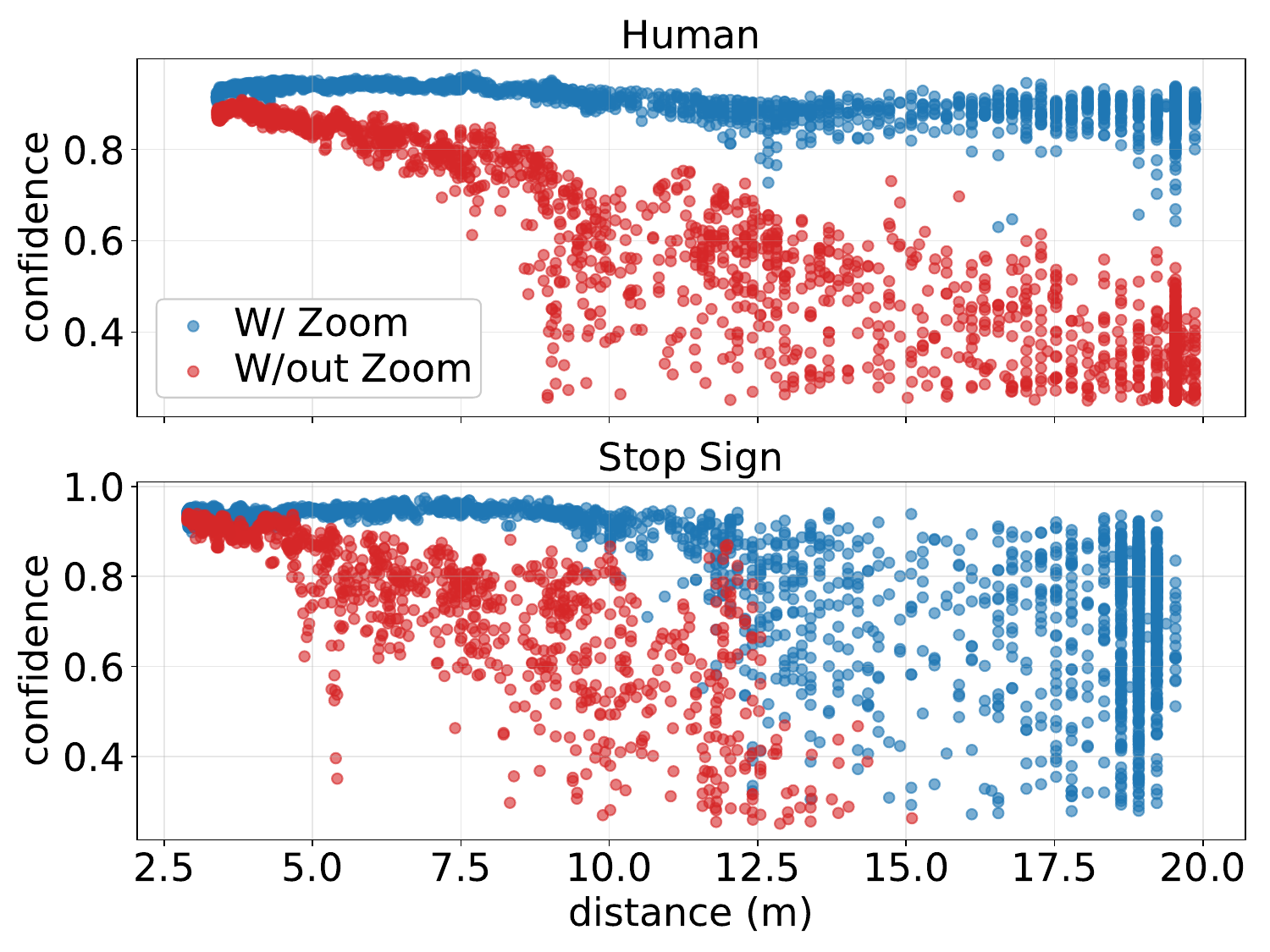}
    \caption{
    \textbf{Effect of Adaptive Zooming on Detection Confidence vs. Distance.}
    Detection confidence for a human and a stop sign is compared with and without the adaptive zoom module. The robot was handheld toward and away from fixed targets in an open indoor space, and the perception pipeline was run in both modes. Without zooming, detection confidence degrades sharply beyond $10$–$12~\si{m}$ due to visual clutter and scale compression. The zooming strategy, by focusing the input on the target and cropping out irrelevant context, maintains high confidence and extends detection range to the camera’s depth limit ($>20~\si{m}$).
    }
    \label{fig:zoom_distance_curves}
\end{figure}

\subsection{Ablation Studies}
To evaluate the contribution of individual components within the NOVA framework, we conduct a series of ablation studies. Each study isolates one module and examines its effect on overall system performance, while keeping the remainder of the stack fixed. The goal is to assess how specific design elements contribute to robustness, accuracy, and safety during target tracking in unstructured environments.

\subsubsection{Adaptive Zoom Strategy}
The adaptive zoom module improves detection robustness by dynamically cropping the image around the expected target location before passing it to the detector. This strategy is designed to address two key challenges: (i) suppressing visually similar distractors that can trigger identity switches, and (ii) preserving the detector’s ability to recognize small, distant targets by maintaining spatial resolution over the region of interest.

\textbf{Multi-Target Robustness.}
We first assess the role of adaptive zoom in scenes with multiple visually similar objects. The setup includes three mannequins placed at different positions, all resembling each other in size and appearance. The UAV is instructed to track one specific mannequin while the others remain in view. 

Figure~\ref{fig:zoom_overlay} shows representative trials. Without zoom, the detector occasionally confuses the target with a nearby distractor, leading to identity switches or temporary tracking failure. With adaptive zooming enabled, the system consistently maintains the correct target across all runs. This supports the hypothesis that reducing visual clutter at the detector input helps identify the target in ambiguous scenes.

\textbf{Long-Range Detection.}
We further evaluate the zoom module's impact on detection reliability at extended distances. Two static targets, a human and a stop sign, are placed in an open indoor space. The robot is handheld and moved toward and away from each target, while the onboard perception stack is run twice: once with zooming enabled and once with full-resolution images passed directly to the detector. Detection confidence is recorded for each case.

Figure~\ref{fig:zoom_distance_curves} summarizes the results. Without zoom, the detector's confidence drops rapidly beyond $10$–$12~\si{m}$, often failing to register the target altogether. In contrast, the zoom module enables consistent detections up to the camera’s maximum effective range of $20$–$25~\si{m}$. By cropping out irrelevant regions and rescaling the image around the target, zooming mitigates scale compression and distractor interference, boosting confidence and range. 

Together, these findings highlight that adaptive zoom contributes to robust tracking in two key regimes: it improves resilience to distractors in multi-object environments and extends detection range under scale-constrained conditions.

\begin{figure}[t]
    \centering
    \includegraphics[width=\linewidth, trim=0 130 420 0, clip]{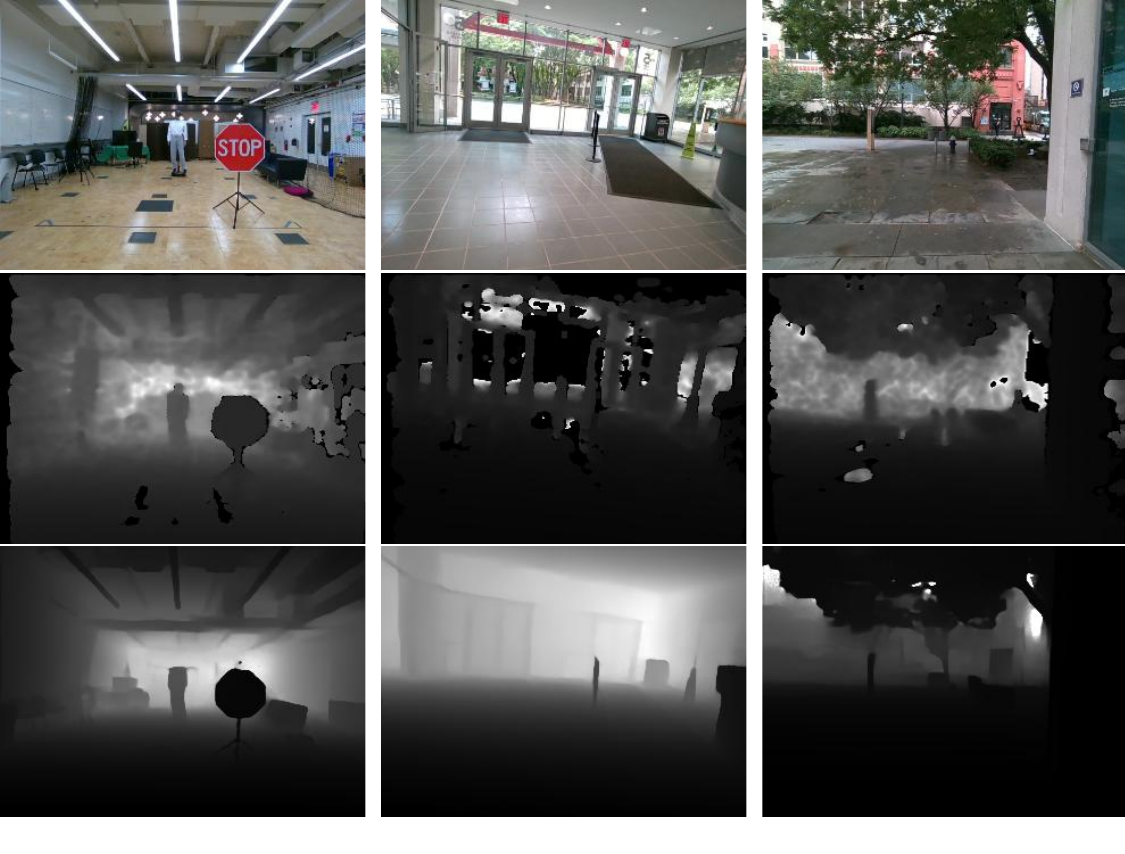}
    \caption{
    \textbf{Qualitative Results of Depth Completion.}
    Top: RGB inputs. Middle: Raw stereo depth from the onboard sensor, which suffers from missing or noisy regions, especially around thin structures, glass, and foliage. Bottom: Completed depth maps produced by our fusion module. The system recovers fine details, such as railings and branches, that are critical for collision avoidance but often missed by stereo matching alone. This enables safe navigation in visually complex environments.
    }
    \label{fig:depth_completion_qual}
\end{figure}

\subsubsection{Depth Completion}
The raw stereo depth output from the robot’s onboard sensor is often sparse and unreliable, particularly in low-texture regions, near thin structures, or under degraded lighting. These limitations reduce the effectiveness of downstream obstacle avoidance and planning components, especially in fast or cluttered environments.

To address this challenge, NOVA incorporates a disparity-based depth completion module. It leverages monocular priors and disparity cues to fill in missing or noisy depth regions, producing denser and smoother maps. Figure~\ref{fig:depth_completion_qual} presents qualitative comparisons between the raw stereo maps and the completed depth output across several representative scenarios. The completed maps better preserve obstacle geometry, enable earlier obstacle detection, and improve motion planning and control safety.

\begin{figure}[t]
    \centering
    \includegraphics[width=\linewidth, trim=0 0 0 0, clip]{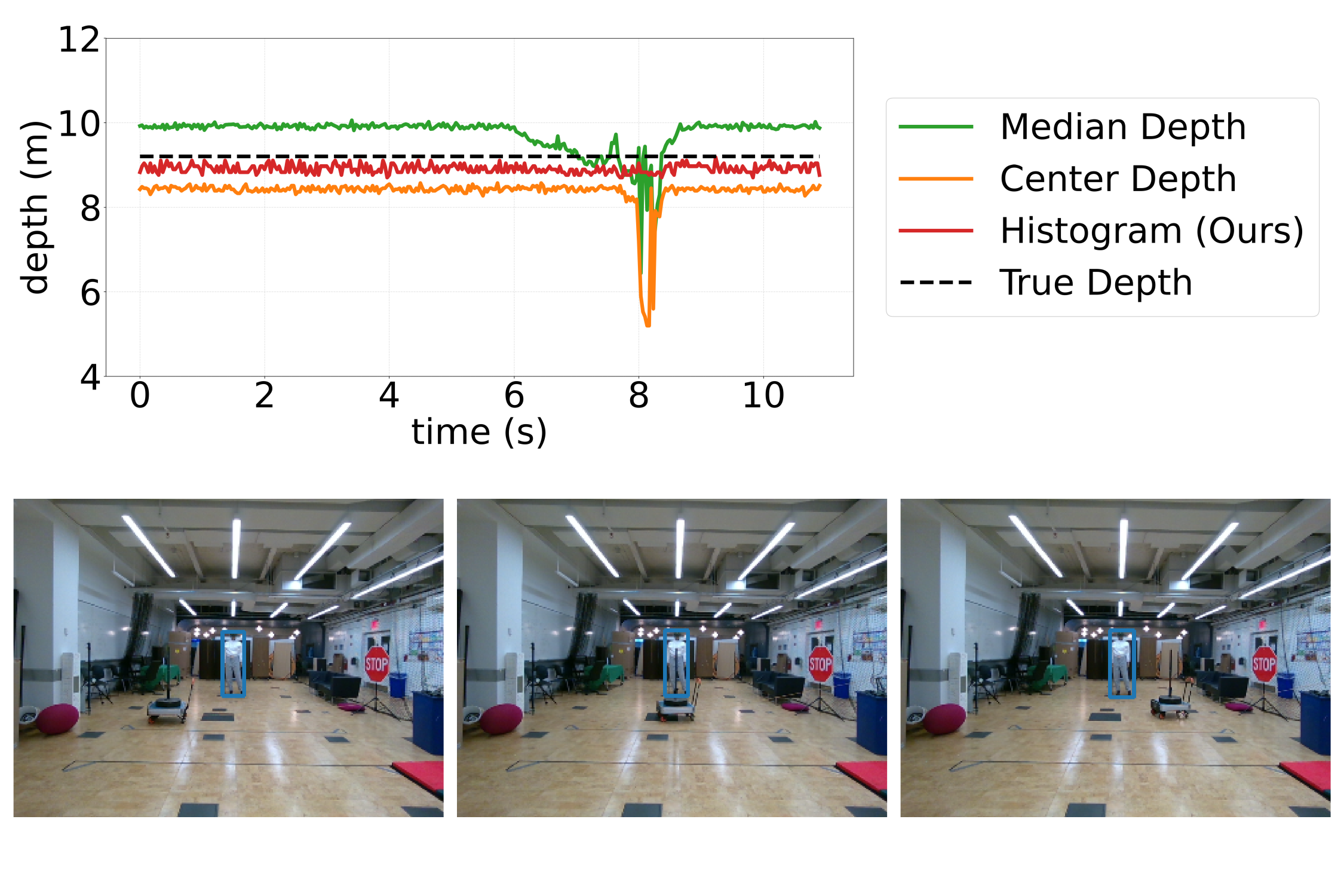}
    \caption{
    \textbf{Robustness of Depth Estimation Methods During Occlusion.}
    A cart with an obstacle occludes the target mid-experiment, inducing background intrusion in the bounding box. The plot shows depth estimates over time using three strategies: center-pixel, mean-pixel, and our histogram-based mode filtering. Ground truth is manually measured. The mean estimator exhibits large spikes during occlusion, and the center-pixel fails intermittently. The histogram-based method remains stable and close to ground truth throughout.
    }
    \label{fig:depth_variance_plot}
\end{figure}

\subsubsection{Histogram-Based Mode Filtering}
Accurate target localization requires reliable estimation of depth within the predicted bounding box. However, raw depth values are often noisy and may include background clutter, particularly in dynamic scenes with partial occlusion. To address this, we introduce a histogram-based mode filtering approach that selects the most frequent depth bin within the bounding box, offering a robust estimate that is resilient to outliers and background interference.

To evaluate this method, we conduct an experiment where the robot hovers in front of a stationary target. A cart carrying a vertical pole is then moved horizontally between the robot and the target, creating a temporary occlusion. During the sequence, we compare three depth estimation strategies: reading the depth value at the center pixel of the bounding box, averaging all valid depth pixels within the box, and applying our proposed histogram-mode filtering.

The ground truth distance to the target is measured manually for reference. As shown in Figure~\ref{fig:depth_variance_plot}, the center-pixel approach produces unstable estimates during occlusion, while the mean-pixel method exhibits a consistent bias toward the background. In contrast, the histogram-based strategy maintains a stable and accurate estimate throughout the occlusion event, closely matching the ground truth.

This improved robustness in depth estimation directly enhances control performance, enabling more stable tracking under visual uncertainty and in cluttered environments.

\section{LIMITATIONS AND FUTURE WORKS}
NOVA operates under two key assumptions that currently define its functional scope and deployment regime.

First, the system assumes that the target is initially visible within the field of view of the onboard camera and belongs to a known object category supported by the detector. This assumption is valid in prompted missions or pre-configured tracking scenarios, but limits applicability in open-world settings where targets may enter the scene later, from arbitrary directions, or lack a pre-defined semantic label. Addressing this constraint would require extending the system to handle open-set detection and tracking, where targets are not assumed to belong to a fixed category and may be specified through visual prompts, user feedback, or learned embeddings~\cite{maalouf2024follow, liu2022opening}. Recent approaches in category-agnostic tracking and self-supervised object discovery offer promising directions for relaxing this assumption and enabling more flexible engagement with previously unseen targets~\cite{saviolo2024unifying}.

Second, the system relies on the ability to estimate the target's depth using stereo-based depth completion. While this approach is sufficient at close to mid-range (typically up to $30$–$40~\si{m}$), it breaks down when the target is detected at longer distances and stereo cues become unreliable. In these conditions, detection may still succeed due to the zoom module, but the depth estimates become noisy or flat, making the x-axis velocity (change in relative depth) unobservable. As a result, the UAV may accelerate rapidly toward the target with limited feedback. In flight experiments, the robot eventually stabilizes once the depth becomes consistent, but the initial approach is often jerky and fast, reflecting the lack of velocity feedback in the unobservable regime.

To relax the reliance on stereo depth, several alternative strategies could be explored. One approach is to augment the sensing stack with additional depth-aware sensors, such as radar or lightweight range finders, which are less sensitive to texture and lighting and can extend the operational range~\cite{nissov2024degradation, kim2024autonomous, zhang2017low} . Another is to infer short-term relative motion using optical flow, either from dense image features or from bounding box displacements of detected objects, including in open-set configurations~\cite{alfarano2024estimating}. A third direction involves registering successive point clouds generated from depth completion, even if sparse or noisy, to recover frame-to-frame translation~\cite{lyu2024rigid}. These approaches would allow for estimation of instantaneous robot velocity without relying on world-frame position or persistent maps, and could remain robust under partial feature disruption due to their minimal or stateless memory requirements.

These assumptions are not intrinsic limitations of the overall framework, but they define the boundaries of the current implementation. Future work targeting open-set object understanding and depth-agnostic motion estimation could expand NOVA’s applicability to more uncertain, long-range, or semantically unconstrained tracking scenarios.

\section{CONCLUSION}
We presented NOVA, an onboard visual-inertial framework for agile target tracking in unstructured and GPS-denied environments. The system avoids reliance on external localization, global maps, or precomputed scene priors. Instead, it formulates perception, estimation, and control directly in the target's reference frame, using only stereo vision and inertial sensing. The approach combines a lightweight detector with adaptive zooming, depth completion, and visual-inertial state estimation, followed by a NMPC that operates under collision-aware constraints derived from onboard sensing.

We validated NOVA across a series of real-world trials that stress the system across different terrain types, motion regimes, and sensory conditions. The system demonstrated consistent performance in forested, urban, and mixed indoor–outdoor environments, maintaining visual lock, respecting safety distances, and operating without manual tuning. Additional experiments evaluated generalization across viewpoint offsets, repeatability over multiple trials, and robustness under degraded GPS and perceptual ambiguity.

Ablation studies confirmed the contribution of individual components, such as adaptive zooming for long-range detection and identity preservation. The system’s limitations were also analyzed, including assumptions about initial visibility, semantic category constraints, and the reliance on stereo-based depth at range.

Together, these results support the conclusion that robust, real-time target tracking can be achieved using only onboard sensing, even in the absence of external infrastructure or structured environments. Future works will focus on relaxing current assumptions, extending to open-set target representations, and enabling perception-driven control under long-range uncertainty.

\bibliographystyle{IEEEtran}
\bibliography{references}

\end{document}